\documentclass[journal]{IEEEtran}
\ifCLASSINFOpdf
\else
\fi

\usepackage{graphicx}
\usepackage{cite}
\usepackage{picinpar}
\usepackage{amsmath}
\usepackage{url}
\usepackage{flushend}
\usepackage[latin1]{inputenc}
\usepackage{colortbl}
\usepackage{soul}
\usepackage{multirow}
\usepackage{pifont}
\usepackage{color}
\usepackage{alltt}
\usepackage[hidelinks]{hyperref}
\usepackage{enumerate}
\usepackage{siunitx}
\usepackage{breakurl}
\usepackage{epstopdf}
\usepackage{pbox}

\usepackage{amssymb}
\usepackage{stfloats}
\usepackage{float}

\usepackage{xcolor}
\usepackage[ruled,linesnumbered]{algorithm2e}
\usepackage{tabularx}
\usepackage{multirow}
\newcommand{\tabincell}[2]{\begin{tabular}{@{}#1@{}}#2\end{tabular}}

\DeclareSymbolFont{largesymbol}{OMX}{yhex}{m}{n}
\DeclareMathAccent{\Widehat}{\mathord}{largesymbol}{"62}

\begin{document}
%
\title{Adaptive Simultaneous Magnetic Actuation and Localization for WCE in a Tubular Environment}
%
%
%

\author{Yangxin~Xu$^{\star}$,~\IEEEmembership{Student~Member,~IEEE,}
        Keyu~Li$^{\star}$,~\IEEEmembership{Student~Member,~IEEE,}\\
        Ziqi~Zhao,
        and~Max~Q.-H.~Meng$^{\sharp}$,~\IEEEmembership{Fellow,~IEEE}
\thanks{This work is partially supported by National Key R \& D program of China with Grant No. 2019YFB1312400, Hong Kong RGC CRF grant C4063-18G and Shenzhen Science and Technology Innovation projects JCYJ20170413161503220 awarded to Max Q.-H. Meng.}
\thanks{Y. Xu and K. Li are with the Department of Electronic Engineering, the Chinese University of Hong Kong, Hong Kong SAR, China (e-mail: yxxu@link.cuhk.edu.hk; kyli@link.cuhk.edu.hk).}
\thanks{Z. Zhao is with the Department of Electronic and Electrical Engineering, the Southern University of Science and Technology, Shenzhen, China (e-mail: zzq2694@163.com).}
\thanks{Max Q.-H. Meng is with the Department of Electronic and Electrical Engineering of the Southern University of Science and Technology in Shenzhen, China, on leave from the Department of Electronic Engineering, the Chinese University of Hong Kong, Hong Kong SAR, China, and also with the Shenzhen Research Institute of the Chinese University of Hong Kong, Shenzhen, China (e-mail: max.meng@ieee.org).}
\thanks{$^{\star}$ indicates equal contribution.}
\thanks{$^{\sharp}$ Corresponding author.}

}

\maketitle

\begin{abstract}
Simultaneous Magnetic Actuation and Localization (SMAL) is a promising technology for active wireless capsule endoscopy (WCE). In this paper, an adaptive SMAL system is presented to efficiently propel and precisely locate a capsule in a tubular environment with complex shapes.
In order to track the capsule with high localization accuracy and update frequency in a large workspace, we propose a mechanism that can automatically activate a sub-array of sensors with the optimal layout during the capsule movement. The improved multiple objects tracking (IMOT) method is simplified and adapted to our system to estimate the 6-D pose of the capsule in real time.
Also, we study the locomotion of a magnetically actuated capsule in a tubular environment, and formulate a method to adaptively adjust the pose of the actuator to improve the propulsion efficiency.
Our presented methods are applicable to other permanent magnet-based SMAL systems, and help to improve the actuation efficiency of active WCE. 
We verify the effectiveness of our proposed system in extensive experiments on phantoms and ex-vivo animal organs. The results demonstrate that our system can achieve convincing performance compared with the state-of-the-art ones in terms of actuation efficiency, workspace size, robustness, localization accuracy and update frequency.
\end{abstract}


%
\IEEEpeerreviewmaketitle

\section{Introduction}

\begin{figure*}[t]
\centering
\includegraphics[scale=1.0,angle=0,width=0.9\textwidth]{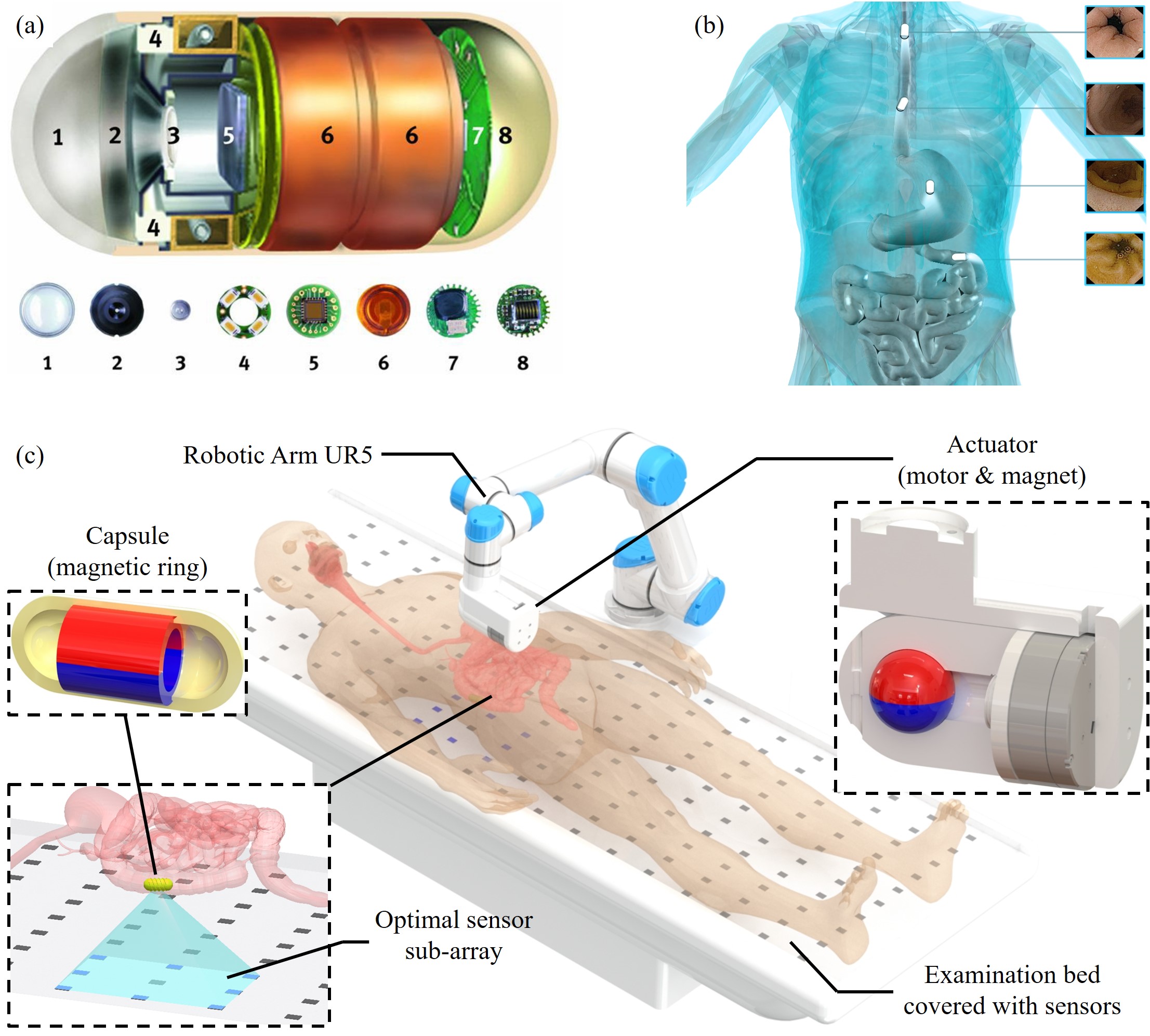}
\caption{(a) The first commercial WCE product, M2A, contains 1) optical dome, 2) lens holder, 3) lens, 4) LEDs, 5) CMOS sensor, 6) batteries, 7) transmitter and 8) antenna. (b) After being swallowed by the patient, WCE can inspect the GI tract in a minimally invasive manner. (c) The overall design of the proposed adaptive SMAL system and its application scenario. The patient is required to swallow a capsule and lie on an examination bed covered with a large sensor array. An actuator consisting of a motor and a spherical magnet is mounted at the end-effector of a robotic arm next to the bed. The actuator rotates above the capsule which contains a magnetic ring to propel it with an adaptive strategy, and the capsule is tracked in real time by an adaptively activated optimal sub-array of sensors.}
\label{Fig_system3D}
\end{figure*}

\IEEEPARstart{A}{ccording} to epidemiologic data, about $8$ million people die from Gastrointestinal (GI) diseases worldwide every year \cite{chan2019gastrointestinal}. Endoscopy plays an important role in the standard procedures of GI diagnosis as it can provide direct visualization of the GI tract.
Since the first Wireless Capsule Endoscopy (WCE) was invented in 2000 (see Fig. \ref{Fig_system3D} (a)), it has been considered as a painless and non-invasive tool for inspection of the human gastrointestinal (GI) tract \cite{iddan2000wireless}. During the diagnosis, the patient only needs to swallow a capsule, then the photos taken by the capsule's camera in the body will be sent out and viewed by the doctor (see Fig. \ref{Fig_system3D} (b)), and the whole process usually takes around $8 \sim 24$ hours \cite{iddan2000wireless}\cite{meng2004wireless}. Active WCE is the concept of endowing the capsule with active locomotion, which has the potential to improve the accuracy of diagnosis and even perform theraputic and surgical procedures. \cite{ciuti2011capsule}. In recent years, simultaneous magnetic actuation \& localization (SMAL) methods for active WCE have been explored to propel and locate the capsule in the intestine, in order to reduce the inspection time and precisely locate the suspected lesions \cite{meng2004wireless}. These methods can be divided into coil-based and permanent magnet-based ones, in terms of the device that generates the magnetic fields for actuation. Compared with the coil-based methods, permanent magnet-based methods generally have the advantages of being more compact, affordable and energy-efficient, and usually have a larger workspace \cite{pittiglio2019magnetic}.

Numerous permanent magnet-based SMAL studies have been reported in the past few decades \cite{abbott2020magnetic}\cite{mateen2017localization}\cite{shamsudhin2017magnetically}. In general, a capsule with embedded magnet(s) is actuated by an external permanent magnet, and their magnetic fields are measured by magnetic sensors \cite{sun2018passive}. The 5-D pose (3-D position, 2-D orientation) of a magnet can be solved from the measured magnetic fields by the magnetic dipole model \cite{cheng1989field}\cite{hu2005efficient}\cite{xu2018free}. The biggest challenge for such systems lies in that the use of magnetic sources for both actuation and localization purposes would result in undesired interferences to the localization system\cite{bianchi2019localization}.
In the most current state-of-the-art SMAL systems \cite{taddese2018enhanced}\cite{popek2017first}, a special structure containing a magnet and several sensors embedded in the capsule was designed so that the sensors only measure the magnetic field of the external actuating magnet \cite{popek2017six}, thereby eliminating the localization interferences.
In \cite{taddese2018enhanced}, the authors used a permanent magnet fixed with an electromagnet as the actuator. They developed Jacobian-based methods for closed-loop velocity control \cite{taddese2016nonholonomic} and localization of the capsule \cite{di2016jacobian}. Their work has been extended to other applications such as ultrasound imaging \cite{norton2019intelligent} and magnetic levitation \cite{barducci2019adaptive}.
The aforementioned methods all utilized the magnetic force for actuation, which is less efficient compared with the magnetic torque, because the torque loss is less prominent with respect to the distance between the capsule and the actuator, compared to the loss in coupling force \cite{mahoney2014generating}. To this end, Popek et al. \cite{popek2017first} used a spherical-magnet actuator \cite{wright2015spherical} to realize rotating magnetic actuation for helical propulsion of a capsule in a lumen \cite{abbott2009should}. During the propulsion, the state of the capsule is distinguished from ``Stuck", ``Synchronous" and ``Step-out" to adopt corresponding actuation strategies \cite{miller2012proprioceptive}\cite{popek20156}.

However, in the aforementioned systems, the special structure inside the capsule occupies a lot of inner space and increases the power consumption. 
Xu et al. \cite{xu2020novelsystem} proposed a SMAL system using rotating actuation and an external sensor array. They used an integral-filter based method \cite{xu2019towards} to eliminate the localization interference, and the capsule's pose is solved by a multi-model localization algorithm after the state of the capsule is automatically detected \cite{xu2020novel}. The use of the external sensors can make the capsule more compact and reduce the power consumption of the capsule. However, the localization method of the system took at least one actuator revolution to update the pose of the capsule, resulting in a low update frequency ($0.5\sim1Hz$).
Based on the multiple magnetic objects tracking (MOT) method \cite{hu2010new}, Song et al.\cite{song2016real} solved the 5-D poses of two magnets with an update frequency of $7Hz$. Xu et al. \cite{xu2020improved} improved the MOT method to estimate the 6-D pose of the capsule from 5-D pose sequences and reached an improved localization update frequency of $25Hz$. Although significant progress has been made in recent works, existing external sensor-based methods are still limited in the following aspects, as reported in \cite{xu2020novelsystem}\cite{xu2020improved}:

\begin{itemize}
\item The confined workspace of the localization system cannot cover the entire abdominal region of the human body. Simply increasing the number of sensors will expand the workspace at the cost of localization update frequency. Therefore, methods for expanding the workspace while maintaining localization accuracy and update frequency should be studied.
\item The robustness of propulsion can be easily affected by the shape of the tubular environment (e.g., straight lumens and sharp bends) due to the varying resistance. Methods that can adaptively change the propulsive force are needed to allow more compliant motion of the capsule in different environments.
\end{itemize}

\begin{figure}[t]
\setlength{\abovecaptionskip}{-0.2cm}
\centering
\includegraphics[scale=1.0,angle=0,width=0.49\textwidth]{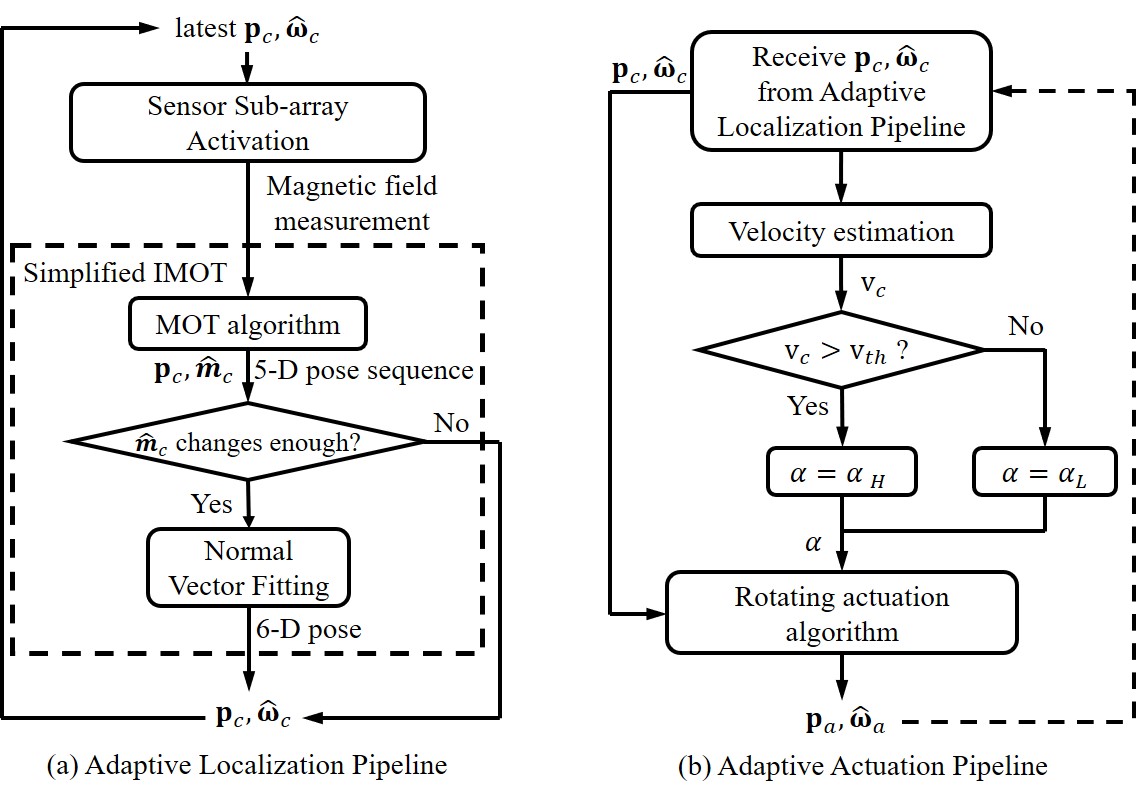}
\caption{The proposed adaptive SMAL workflow includes two parallel pipelines. (a) shows the adaptive localization pipeline. A sub-array of sensors with the optimal layout is selected and activated according to the latest pose $\mathbf{p}_{c}, \widehat{\pmb{\omega}}_{c}$ of the capsule, and the superimposed magnetic field is measured. The pose of the capsule is updated with a simplified version of the IMOT algorithm \cite{xu2020improved}, which utilizes the Normal Vector Fitting (NVF) to estimate the 6-D pose from the 5-D pose sequence given by the MOT algorithm. The pose of the capsule is only updated when $\widehat{\mathbf{m}}_{c}$ changes enough. (b) shows the adaptive actuation pipeline. According to the pose estimation from the first pipeline, the capsule's velocity ${v}_{c}$ is estimated and compared with a threshold $v_{th}$ to determine the actuating angle $\alpha$. Then based on $\mathbf{p}_{c}, \widehat{\pmb{\omega}}_{c}$, the pose of the actuator $\mathbf{p}_{a}, \widehat{\pmb{\omega}}_{a}$ is calculated by the rotating actuation algorithm \cite{mahoney2014generating} to close the loop.}
\label{Fig_workflow}
\end{figure}

To address these issues, we present a novel \textit{adaptive SMAL} system to allow efficient propulsion and accurate localization of a capsule in a tubular environment. The overall design of our system and its application scenario are shown in Fig. \ref{Fig_system3D} (c). The proposed workflow in Fig. \ref{Fig_workflow} includes two parallel processing pipelines. The first pipeline estimates the 6-D pose of the capsule with a high update frequency based on the measurements from an adaptively activated optimal sub-array of sensors. In the second pipeline, an adaptive propulsive force is applied to the capsule by changing the actuating angle based on the current velocity of the capsule.

The main contributions of this paper are three-fold:

\begin{itemize}
\item A novel localization approach based on an adaptively activated sub-array of sensors with the optimal layout is proprosed to track the capsule, which can significantly expand the workspace while increasing the accuracy and update frequency of localization. The 6-D pose of the capsule is solved with an adapted version of the improved multiple objects tracking (IMOT) algorithm \cite{xu2020improved}.
\item The locomotion of the capsule actuated by the rotating magnetic field in a tubular environment with different shapes is studied. On this basis, an adaptive strategy that changes the pose of the actuator in real time is developed to improve the actuation efficiency and allow compliant motion of the capsule in different environments.
\item An overall adaptive SMAL framework is presented to efficiently propel and precisely locate a magnetically actuated capsule in a tubular environment with complex shapes, and the system is validated in extensive experiments on phantoms and ex-vivo animal organs.
\end{itemize}

The remainder of this paper is organized as follows. Section II introduces the nomenclature of this paper. Details of the adaptive localization and actuation methods are introduced in Section III and Section IV, respectively. Experimental results are presented and discussed in Section V, before we summarize our work and conclude this article in Section VI.

\section{NOMENCLATURE}

Throughout this paper, lowercase normal fonts denote scalars (e.g., $\mu_{0}$), lowercase bold fonts denote vectors (e.g., $\textbf{b}$), and uppercase bold fonts denote matrices (e.g. $\textbf{M}$). The vector with a ``hat'' symbol indicates that the vector is a unit vector of the original vector (e.g., $\widehat{\textbf{r}}$ is the unit vector of $\textbf{r}$). $\textbf{I}_{n}$ denotes $n \times n$ identity matrix. $Rot_{\mathbf{a}}(\theta)$ refers to the rotation of $\theta$ degrees around a vector $\mathbf{a}$, and $Rot_{k}(\theta)$ refers to the rotation of $\theta$ degrees around +$k$-axis, $k \in \{x,y,z\}$.

\section{Activation of an optimal sensor sub-array and capsule localization with simplified IMOT}

\subsection{Selection and activation of an optimal sensor sub-array}

Instead of adopting a fixed $4\times4$ sensor array as in \cite{xu2020improved}, we propose in this section a method to adaptively activate a sub-array of $N$ sensors with the optimal layout from an original array of $N_{x} \times N_{y}$ sensors covering the entire workspace.

\begin{figure}[t]
\setlength{\abovecaptionskip}{-0.0cm}
\centering
\includegraphics[scale=1.0,angle=0,width=0.42\textwidth]{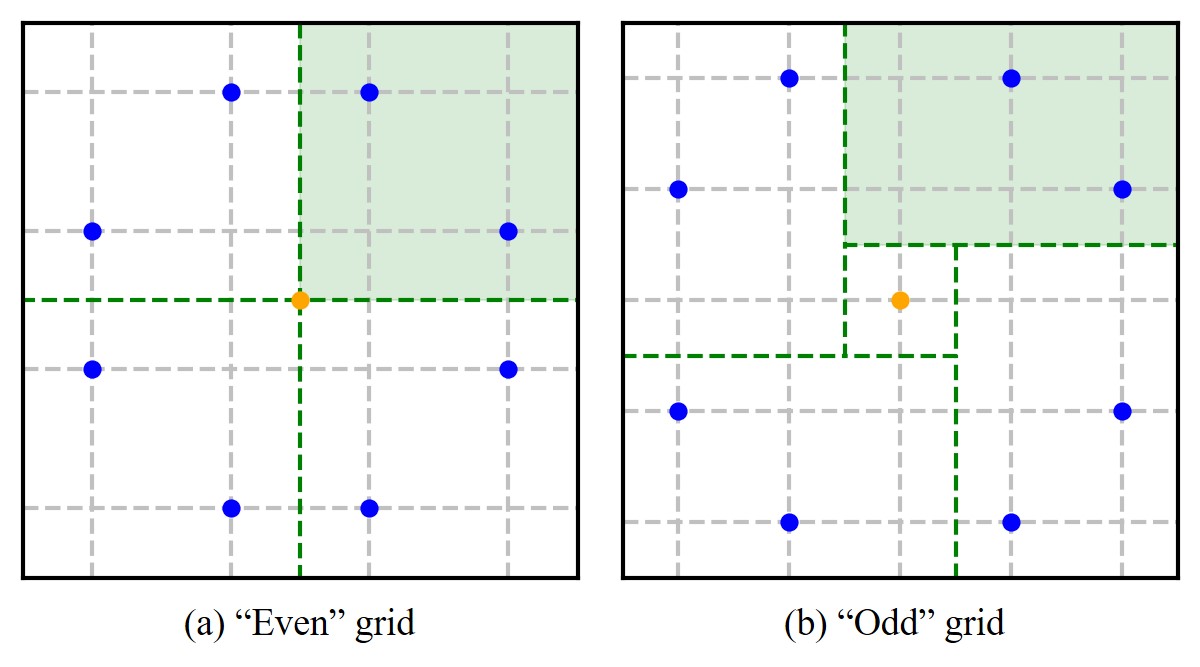}
\caption{Example layouts of $8$ sensors in an (a) ``$4\times4$" and an (b) ``$5\times5$" grid. ``$4\times4$" and ``$5\times5$" are determined by the number of sensors that can be placed in each row (column) of the grid. Since we assume the layout satisfies $4$-fold rotational symmetry about the center point (yellow), only one quarter of the grid (green shaded region) needs to be considered in the layout design. For example, $2$ sensors (blue points) are arranged in the quarter and copied to the other $3$ quarters by rotating $90^{\circ}$, $180^{\circ}$, and $270^{\circ}$ around the center (yellow point).}
\label{Fig_layouts}
\end{figure}

First, we study the optimal layout of a fixed number of sensors to maximize the localization accuracy. Inspired by \cite{song2017design}, we utilize the combinatorial mathematics based method to list all possible layouts of $N=8$ and $9$ sensors in a grid subject to the assumption that the layouts satisfy $4$-fold rotational symmetry around the center. Depending on the number of sensors that can be placed in each row (column), the grid is called ``$4\times4$" or ``$5\times5$". We show two example layouts of $8$ sensors in Fig. \ref{Fig_layouts}. Since a sparser layout will result in a decrease in localization accuracy, we ignore grids larger than that in Fig. \ref{Fig_layouts}. Due to the requirement of symmetry, only one quarter of the grid needs to be considered in the layout design. For $N=8$, two sensors should be placed in one quarter in both grids, forming a total of $\mathbf{C}_{4}^{2}+\mathbf{C}_{6}^{2}=21$ layouts. For $N=9$, two sensors should be placed in one quarter of the ``$5\times5$" grid and one sensor should be placed at the center, forming a total of $\mathbf{C}_{6}^{2}=15$ layouts. Therefore, a total of $36$ layouts for the sensor sub-array are proposed as the candidates.

\begin{figure}[t]
\setlength{\abovecaptionskip}{-0.4cm}
\centering
\includegraphics[scale=1.0,angle=0,width=0.49\textwidth]{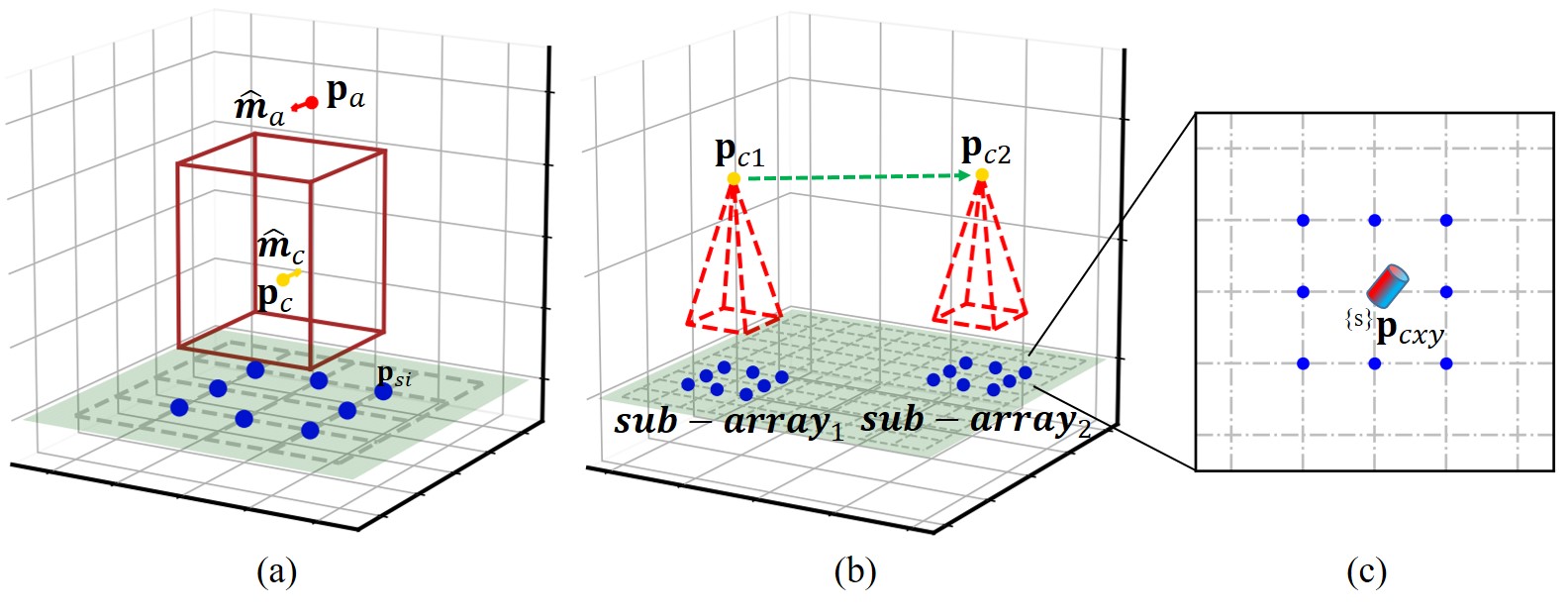}
\caption{(a) In the localization accuracy test for different layouts, the capsule is randomly placed inside a box-shape region $5 cm$ above the sensors, with a bottom area equal to the size of the layout and a height of $15 cm$. The actuator's pose ($\mathbf{p}_{a},\widehat{\mathbf{m}}_{a}$) and the capsule's pose ($\mathbf{p}_{c},\widehat{\mathbf{m}}_{c}$) are depicted in red and yellow, respectively. (b) During propulsion of the capsule from $\mathbf{p}_{c1}$ to $\mathbf{p}_{c2}$, the optimal sensor sub-array is selected and activated in real time. (c) shows the top view of the capsule and the optimal sensor sub-array.}
\label{Fig_TestTracking}
\end{figure}

The localization accuracy of the sensor sub-array with different layouts is tested in simulation. As shown in Fig. \ref{Fig_TestTracking} (a), the capsule is randomly placed inside a box-shape region above the sensors, and the pose of the actuator is determined by the rotating actuation algorithm \cite{mahoney2014generating}. $\mathbf{p}_{a}=(p_{ax},p_{ay},p_{az})$, $\mathbf{p}_{c}=(p_{cx},p_{cy},p_{cz})$ $\in\mathbb{R}^{3\times1}$ refer to the positions of the actuator and capsule, respectively, and $\widehat{\mathbf{m}}_{a}$, $\widehat{\mathbf{m}}_{c}$ $\in\mathbb{R}^{3\times1}$ are corresponding unit magnetic moments. $\mathbf{p}_{si}\in\mathbb{R}^{3\times1}$ represents the position of the $i$-th sensor, and the simulated superimposed magnetic field measured by the $i$-th sensor is denoted by $\mathbf{b}_{si}$. According to the MOT algorithm \cite{hu2010new}, the 5-D poses of the actuator and the capsule ($\mathbf{p}_{a},\widehat{\mathbf{m}}_{a},\mathbf{p}_{c},\widehat{\mathbf{m}}_{c}$) can be solved by:

\begin{equation}
\label{F_MOT}
\begin{aligned}
\min_{\mathbf{p}_{a},\widehat{\mathbf{m}}_{a},\mathbf{p}_{c},\widehat{\mathbf{m}}_{c}} \quad & \sum_{i=1}^{N}  \|\mathbf{b}_{ai}(\mathbf{r}_{ai},\widehat{\mathbf{m}}_{a}) + \mathbf{b}_{ci}(\mathbf{r}_{ci},\widehat{\mathbf{m}}_{c})-\mathbf{b}_{si}\| \\
\textrm{subject to} \quad & \|\widehat{\mathbf{m}}_{a}\|=1, \|\widehat{\mathbf{m}}_{c}\|=1\\
\end{aligned}
\end{equation}

\noindent where $\mathbf{r}_{ai}=\mathbf{p}_{si}-\mathbf{p}_{a}$ and $\mathbf{r}_{ci}=\mathbf{p}_{si}-\mathbf{p}_{c}$ are vectors pointing from each magnet's position $\mathbf{p}_{a}$, $\mathbf{p}_{c}$ to the $i$-th sensor position $\mathbf{p}_{si}$. The theoretical magnetic fields $\mathbf{b}_{ai}$, $\mathbf{b}_{ci}$ $\in\mathbb{R}^{3\times1}$ at $\mathbf{p}_{si}$ can be calculated with the magnetic dipole model \cite{cheng1989field}.

\begin{figure}[t]
\setlength{\abovecaptionskip}{-0.0cm}
\centering
\includegraphics[scale=1.0,angle=0,width=0.48\textwidth]{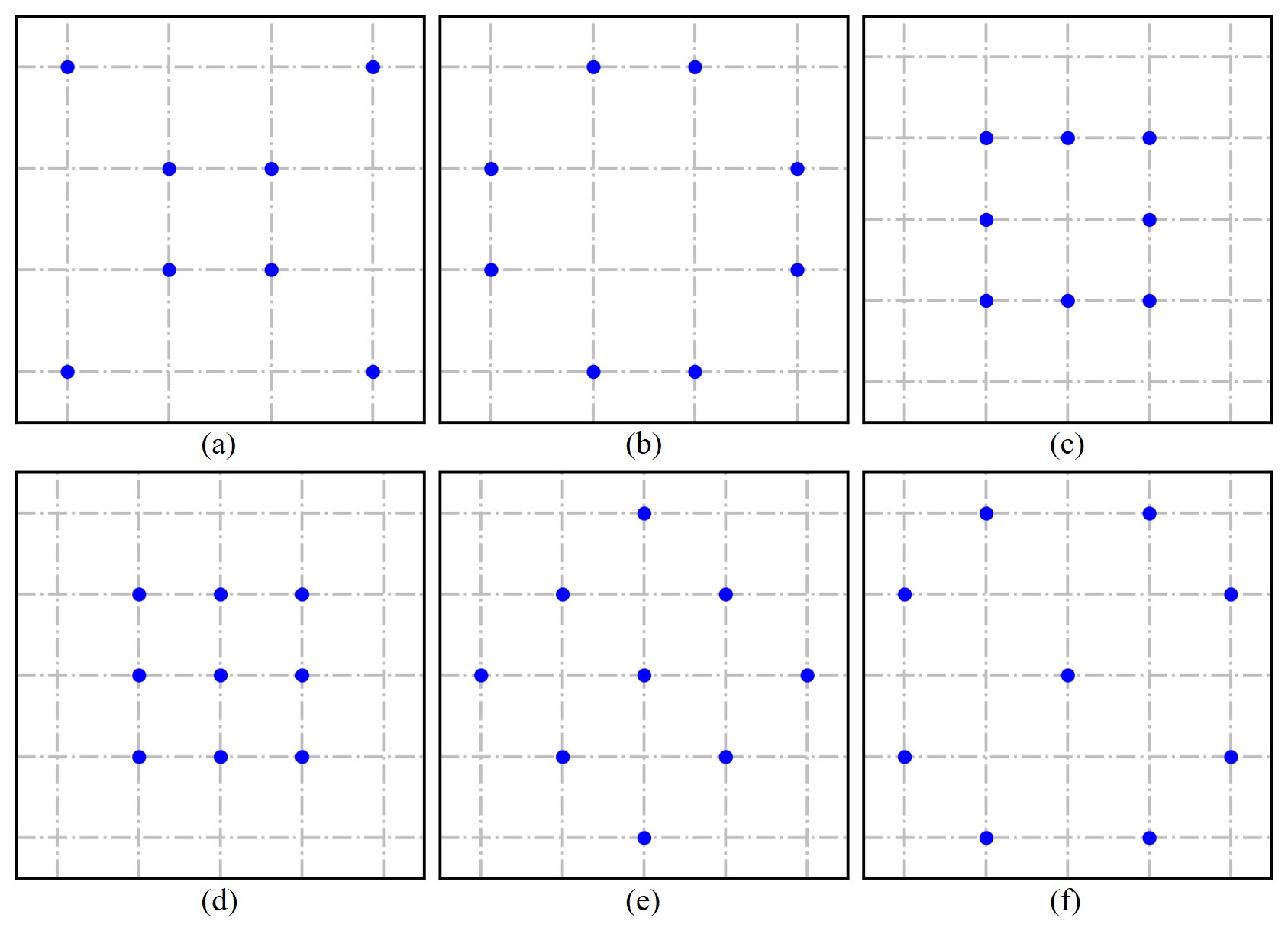}
\caption{The first $6$ layouts with the highest localization accuracy among a total of $36$ layouts are shown, and (c) is preliminarily determined as the ``optimal layout'' from the simulation result.}
\label{Fig_layouts_candidates}
\end{figure}

In order to determine the optimal layout for the sensor sub-array, we illustrate the six best layouts in terms of localization accuracy in simulation among a total of $36$ proposed layout candidates, as shown in Fig. \ref{Fig_layouts_candidates}, and Table \ref{T_layoutsimulation} presents the quantitative results for a comparison. Based on the simulation results, we select layout (c) as the optimal layout, which achieves the highest accuracy of $4 mm$ / $4.4^{\circ}$ in position and orientation, respectively.

\begin{table}[b] \footnotesize
\centering
\caption{Localization accuracy with different layouts in simulation}
\begin{tabular}{|p{2.5cm}<{\centering}|p{2.5cm}<{\centering}|p{2.5cm}<{\centering}|}
\hline
\multirow{2}{*}{\tabincell{c}{Sensor Array Layout}} & \multirow{2}{*}{\tabincell{c}{Position Error}} & \multirow{2}{*}{\tabincell{c}{Orientation Error}} \\
{} & {} & {} \\
\hline
{(a)} & {$4.7 mm$} & {$4.5^{\circ}$} \\
\hline
{(b)} & {$6.3 mm$} & {$5.7^{\circ}$} \\
\hline
{(c)} & {$\mathbf{4.0 mm}$} & {$\mathbf{4.4^{\circ}}$} \\
\hline
{(d)} & {$4.4 mm$} & {$4.6^{\circ}$} \\
\hline
{(e)} & {$5.9 mm$} & {$6.0^{\circ}$} \\
\hline
{(f)} & {$5.8 mm$} & {$5.9^{\circ}$} \\
\hline
\end{tabular}
\label{T_layoutsimulation}
\end{table}

As shown in Fig. \ref{Fig_TestTracking} (b), during the propulsion of the capsule, we activate the optimal sensor sub-array below the capsule in real time. The center index of the sub-array ($c_{x},c_{y}$) is determined by:

\begin{equation}
\label{F_labels}
\begin{aligned}
&c_{k}=
\left\{
\begin{aligned}
1 &,\ round(\frac{^{\{S\}}{p}_{ck}}{D}) < 1 \\
N_k-2 &,\ round(\frac{^{\{S\}}{p}_{ck}}{D})  > N_k-2 \\
round(\frac{^{\{S\}}{p}_{ck}}{D})  &, \  otherwise \\
\end{aligned}
\right.\\
& k \in \{x, y\}
\end{aligned}
\end{equation}

\noindent where $^{\{S\}}\mathbf{p}_{cxy}=(^{\{S\}}{p}_{cx},^{\{S\}}{p}_{cy})$ is the XY position of the capsule $w.r.t.$ the sensor frame $\{S\}$, $round(\cdot)$ is the rounding function, and $D$ is the distance between sensors in the original sensor array. After the center index ($c_{x},c_{y}$) is determined, we can get the index of each sensor in the sub-array by $(i,j) \in \{(i,j)|i \in\{c_{x}-1,c_{x},c_{x}+1\}, j \in\{c_{y}-1,c_{y},c_{y}+1\}, (i,j) \not= \{(c_{x},c_{y})\}\} $.

After the optimal sensor sub-array is activated, an adapted version of the IMOT algorithm will be used to estimate the 6-D pose of the capsule based on the measurements of the superimposed magnetic field, as described in Section~III-B. We will demonstrate in our experiments that this adaptive sensor activation method can achieve accurate localization of the capsule in a large workspace and greatly increase the update frequency.

\subsection{Simplified IMOT for capsule tracking}

\begin{figure}[t]
\setlength{\abovecaptionskip}{-0.1cm}
\centering
\includegraphics[scale=1.0,angle=0,width=0.30\textwidth]{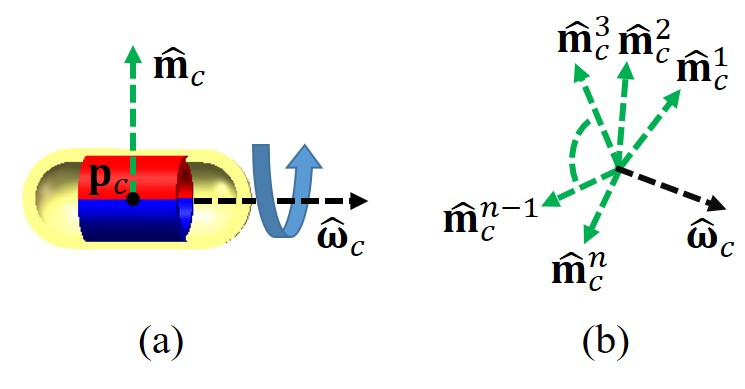}
\caption{(a) Illustration of the 6-D pose of the capsule (3-D position $\mathbf{p}_c$, 2-D unit magnetic moment $\widehat{\mathbf{m}}_{c}$ and the heading direction $\widehat{\pmb{\omega}}_{c}$). (b) Illustration of the Normal Vector Fitting (NVF) method, which can estimate $\widehat{\pmb{\omega}}_{c}$ from a sequence of $n$ magnetic moments ($\widehat{\mathbf{m}}^{1}_{c},\cdots,\widehat{\mathbf{m}}^{n}_{c}$).}
\label{Fig_imot_NVF}
\end{figure}

We follow the definition of the 6-D pose of a rotating capsule in \cite{xu2020improved} as the 3-D position $\mathbf{p}_c$, 2-D unit magnetic moment $\widehat{\mathbf{m}}_{c}$ and the heading direction of the capsule $\widehat{\pmb{\omega}}_{c}$. As shown in Fig. \ref{Fig_imot_NVF} (a), a permanent magnetic ring is embedded inside the capsule, whose magnetic moment $\widehat{\mathbf{m}}_{c}$ is along the radial direction of the capsule. Hence, $\widehat{\mathbf{m}}_{c}$ is always perpendicular to the heading direction of the capsule $\widehat{\pmb{\omega}}_{c}$ due to this installation. Since the position of the capsule has 3 DOFs, the unit vectors $\widehat{\mathbf{m}}_{c}$ and $\widehat{\pmb{\omega}}_{c}$ each has 2 DOFs but are mutually perpendicular, the combination of $\mathbf{p}_c$, $\widehat{\mathbf{m}}_{c}$ and $\widehat{\pmb{\omega}}_{c}$ have a total of 6 DOFs. The 5-D pose of the capsule ($\mathbf{p}_c$ and $\widehat{\mathbf{m}}_{c}$) can be obtained using the multiple magnetic objects tracking (MOT) algorithm \cite{hu2010new}. In our previous work \cite{xu2020improved}, we proposed the IMOT algorithm to estimate the moving direction $\widehat{\pmb{\omega}}_{c}$ based on a sequence of 5-D poses of the capsule using the following three steps, namely, i) Normal Vector Fitting (NVF), ii) B\'{e}zier Curve Gradient (BCG), and iii) Spherical Linear Interpolation (SLI).

The NVF step approximates the heading direction of the rotating capsule as its current rotation axis. As shown in Fig. \ref{Fig_imot_NVF} (b), during the rotation of the capsule, a sequence of $n$ magnetic moments can be obtained by the MOT algorithm as ($\widehat{\mathbf{m}}^{1}_{c},\cdots,\widehat{\mathbf{m}}^{n}_{c}$). Let $\textbf{M}=\left[\begin{matrix}\widehat{\mathbf{m}}_{c}^{1}, \cdots, \widehat{\mathbf{m}}_{c}^{n} \\ \end{matrix}\right]\in\mathbb{R}^{3 \times n}$, and $\widehat{\mathbf{m}}^{i}_{c} \perp \widehat{\pmb{\omega}}_{c}$, $i \in \{1,\cdots,n\}$, so we have $\textbf{M}^{T}\widehat{\pmb{\omega}}_{c}=\mathbf{0}$. Therefore, the heading direction of the capsule $\widehat{\pmb{\omega}}_{c}$ can be estimated by solving the following optimization problem:

\begin{equation}
\label{F_NVF}
\begin{aligned}
\mathop{\arg\min}_{\widehat{\pmb{\omega}}_{c}} \quad & \|\mathbf{M}^{T}\widehat{\pmb{\omega}}_{c}\|\\
\textrm{subject to} \quad & \|\widehat{\pmb{\omega}}_{c}\|=1\\
\end{aligned}
\end{equation}

When the moving speed of the capsule is relatively slow, NVF can locate the capsule with high accuracy. However, as the speed of the capsule increases, the accuracy of the NVF method gradually decreases, mainly because the moving direction of the capsule may have changed during the measurement of $n$ magnetic moments \cite{xu2020improved}.
Therefore, the BCG step is used to locate the capsule with high accuracy when the capsule moves in a higher speed. The Gaussian Mixture Model (GMM) combined with the expectation-maximization (EM) algorithm is used to cluster the points in the trajectory of the capsule over a period of time, and a smooth quadratic B\'{e}zier curve is generated to fit the trajectory. Then the directional derivative at the end of the trajectory is used to estimate the current moving direction of the capsule. The BCG method can achieve higher localization accuracy when the capsule moves with a high speed, but the localization error will become larger when capsule moves slowly \cite{xu2020improved}.
Finally, the SLI step uses Spherical Linear Interpolation to combine the advantages of NVF and BCG to obtain the final localization result of the capsule.
The readers can refer to \cite{xu2020improved} for more details about this method.

\begin{figure}[t]
\setlength{\abovecaptionskip}{-0.4cm}
\centering
\includegraphics[scale=1.0,angle=0,width=0.49\textwidth]{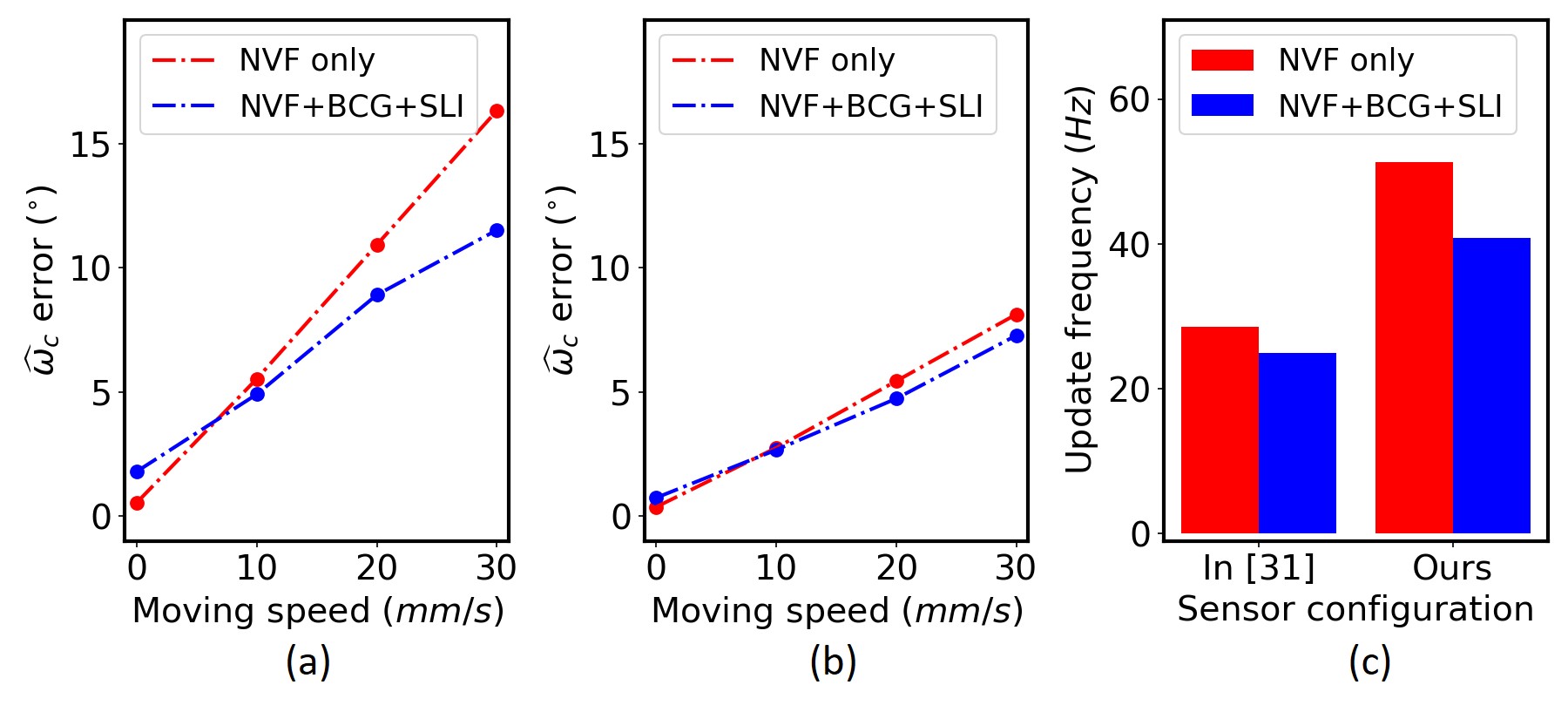}
\caption{Localization accuracy and update frequency of the original IMOT (NVF+BCG+SLI) \cite{xu2020improved} and our simplified version (NVF only) under different sensor configurations. (a) When using the sensor configuration in \cite{xu2020improved}, the estimation error of $\widehat{\pmb{\omega}}_{c}$ by only NVF increases faster than that by NVF+BCG+SLI as the moving speed of the capsule increases. (b) When using the the sensor configuration proposed in this paper, the localization accuracy of only NVF is largely improved, which is only slightly worse than that of NVF+BCG+SLI, but (c) the localization update frequency with our simplified method is significantly increased.}
\label{Fig_imot_new}
\end{figure}

However, through a number of experiments, we found that the running time of the IMOT algorithm is largely limited by the BCG part due to the GMM+EM cluster step. In fact, since we adopt the optimal sensor layout consisting of $8$ sensors in this work while the original IMOT method \cite{xu2020improved} uses $16$ sensors, the speed of reading sensor data and solving the optimization problem in the MOT algorithm are much faster than that in \cite{xu2020improved}. As a result, the time required to obtain $n$ magnetic moments for the NVF method is greatly shortened. Therefore, the moving direction of the capsule can be regarded as constant during this short period. Based on this consideration, we simplify the IMOT algorithm by waiving the BCG and SLI steps, and only use the NVF method to estimate $\widehat{\pmb{\omega}}_{c}$. We empirically show that under our optimal sensor configuration, this simplification can increase the localization update frequency while maintaining the localization accuracy when the capsule moves at different speeds, as shown in Fig. \ref{Fig_imot_new}.

\subsection{Adaptive localization algorithm}

We summarize the adaptive localization algorithm in Algorithm \ref{Alg_loc}, which corresponds to the pipeline in Fig. \ref{Fig_workflow} (a).

\begin{algorithm}[t] 
\caption{Adaptive Localization Algorithm}
\label{Alg_loc}
\KwIn{\textit{5-D pose sequence} of length $n$ $ \{ (\mathbf{p}_{c}^{(t-n)},\widehat{\pmb{m}}_{c}^{(t-n)}),\cdots,(\mathbf{p}_{c}^{(t-1)},\widehat{\pmb{m}}_{c}^{(t-1)}) \}$ \\ \ \ \ \ \ \ \  \ \ and the latest $\widehat{\pmb{\omega}}_{c}^{(t-1)}$}
\KwOut{current pose $(\mathbf{p}_{c}^{(t)}$, $\widehat{\pmb{m}}_{c}^{(t)}$, $\widehat{\pmb{\omega}}_{c}^{(t)})$}
Solve the center index of the sub-array ($c_{x}$,$c_{y}$) by (\ref{F_labels})\;
Activate the sensors in the sub-array with index $(i,j) \in \{(i,j)|i \in\{c_{x}-1,c_{x},c_{x}+1\}, j \in\{c_{y}-1,c_{y},c_{y}+1\}, (i,j) \not= \{(c_{x},c_{y})\}\}$\;
Get the magnetic field measurements $\mathbf{b}_{s}$, $s\in\{1,\cdots,8\}$ from the selected sensors\;
Solve the current 5-D pose of the capsule $(\mathbf{p}_{c}^{(t)}$, $\widehat{\pmb{m}}_{c}^{(t)}$) by (\ref{F_MOT})\;
\For{$i=1,\cdots,n$}
{Update the \textit{5-D pose sequence}
$(\mathbf{p}_{c}^{(t-i)},\widehat{\pmb{m}}_{c}^{(t-i)}) \leftarrow (\mathbf{p}_{c}^{(t-i+1)},\widehat{\pmb{m}}_{c}^{(t-i+1)})$}
\If{$\arccos((\widehat{\mathbf{m}}_{c}^{(t-n)}){}^{T}\widehat{\mathbf{m}}_{c}^{(t-1)})$ is small}
{$\widehat{\pmb{\omega}}_{c}^{(t)}=\widehat{\pmb{\omega}}_{c}^{(t-1)}$\;}
\Else{Update $\widehat{\pmb{\omega}}_{c}^{(t)}$ with the NVF method by (\ref{F_NVF})\;}
\Return{capsule's current pose} $(\mathbf{p}_{c}^{(t)}$, $\widehat{\pmb{m}}_{c}^{(t)}$, $\widehat{\pmb{\omega}}_{c}^{(t)})$\;
\end{algorithm}

\section{Analysis of capsule locomotion in a tubular environment and formulation of adaptive actuation}

\subsection{Analysis of capsule locomotion in a tubular environment}

In this section, we first formulate the rotating magnetic actuation model based on the concept of \textit{actuating angle}, and then analyze the capsule locomotion in a tubular environment with different shapes under different actuating angles, before we introduce an adaptive method to adjust the pose of the actuator for efficient and robust propulsion of the capsule. It is assumed that the inner diameter of the tubular environment is slightly larger than the diameter of the capsule, and the friction and deformation of the tubular environment are negligible. 

\begin{figure*}[t]
\setlength{\abovecaptionskip}{-0.0cm}
\centering
\includegraphics[scale=1.0,angle=0,width=0.98\textwidth]{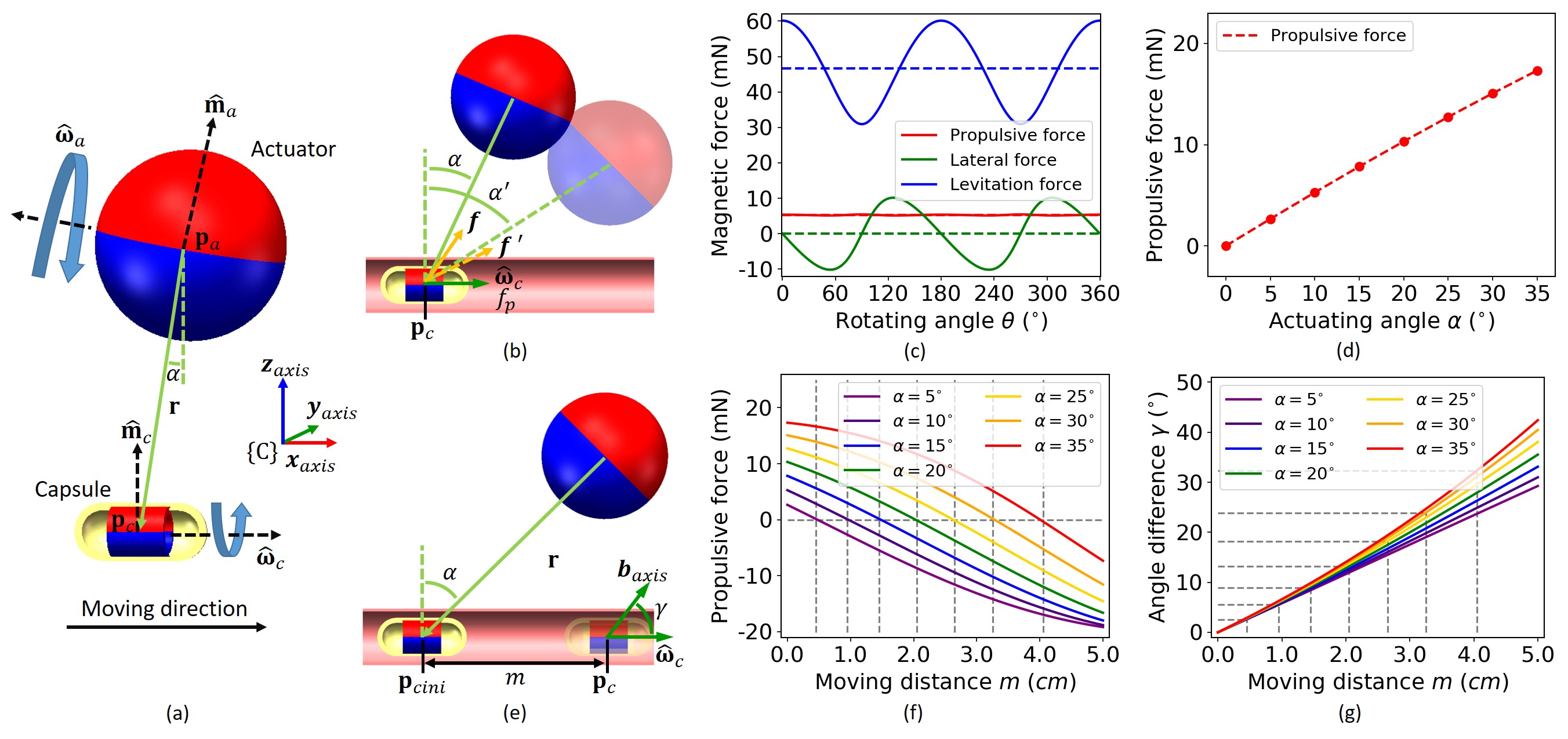}
\caption{(a) The actuator rotates around its heading direction $\widehat{\pmb{\omega}}_{a}$ at position $\mathbf{p}_{a}$ to generate a magnetic field that rotates the capsule around its heading direction $\widehat{\pmb{\omega}}_{c}$ at $\mathbf{p}_{c}$. (b) Different actuating angles $\alpha$ result in different magnetic forces $\mathbf{f}$ applied to the capsule at $\mathbf{p}_{c}, \widehat{\pmb{\omega}}_{c}$. $f_p$ is the component of $\mathbf{f}$ in the direction of $\widehat{\pmb{\omega}_{c}}$. (c) shows the propulsive force, the lateral force and the levitation force applied to the capsule over one actuator revolution when $\alpha=10^{\circ}$, and the dashed lines refer to the corresponding average values. (The red dashed line is obscured by the red solid line.) (d) The propulsive force $f_{p}$ gradually increases with the actuating angle $\alpha$. (e) The capsule moves from $\mathbf{p}_{cini}$ to $\mathbf{p}_{c}$ under the rotating actuation. $\gamma$ is the angle difference between the moving direction of the capsule $\widehat{\pmb{\omega}}_{c}$ and the rotation axis of the magnetic field $\mathbf{b}_{axis}$. (f) The propulsive force $f_p$ decreases with the moving distance of the capsule $m$ under different actuating angles. (g) The angle difference $\gamma$ increases with the moving distance of the capsule $m$ under different actuating angles.}
\label{Fig_fixedmoving}
\end{figure*}

In order to introduce the concept of the \textit{actuating angle}, we first define a capsule-centric reference frame \{C\} as shown in Fig. \ref{Fig_fixedmoving} (a). The origin of the frame is located at the current position of the capsule $\mathbf{p}_c$, and the $x$-axis is always aligned with the heading direction of the capsule, i.e., $^C\widehat{\pmb{\omega}}_{c}=\left(\begin{matrix}1&0&0\end{matrix}\right)^{T}$. The $z$-axis of \{C\} is defined to be located in the vertical plane (the plane formed by the $x$-axis and the vertical line that passes through the origin of \{C\}). Then the $y$-axis is determined using the right-hand rule. The actuator is also located in this vertical plane at position $\mathbf{p}_a$, and the angle between the vector $\overrightarrow{\mathbf{p}_c \mathbf{p}_a}$ and the $z$-axis of \{C\} is defined as the \textit{actuating angle} $\alpha$. Since \{C\} is used as the reference frame in the mathematical derivation in this subsection, the superscript ``C" is omitted in the following for simplicity. Assume that the initial heading direction of the actuator $\widehat{\pmb{\omega}}_{a}$ is aligned with the $x$-axis, i.e., $\widehat{\pmb{\omega}}_{a,ini}=\left(\begin{matrix}1&0&0\end{matrix}\right)^{T}$, the initial magnetic moment of the actuator $\widehat{\mathbf{m}}_{a,ini}$ is defined to be aligned with the $z$-axis, i.e., $\widehat{\mathbf{m}}_{a,ini}=\left(\begin{matrix}0&0&1\end{matrix}\right)^{T}$.

According to the rotating magnetic actuation algorithm \cite{mahoney2014generating}, in order to generate a magnetic field that rotates the capsule around its heading direction $\widehat{\pmb{\omega}}_{c}$ at position $\mathbf{p}_{c}$, the rotation axis of the actuating magnet $\widehat{\pmb{\omega}}_{a}$ can be calculated by:

\begin{equation}
\label{F_rotating_actuation}
\widehat{\pmb{\omega}}_{a}=\Widehat{\left((3\widehat{\mathbf{r}}{\widehat{\mathbf{r}}}^{T}-\mathbf{I}_{3})\ \widehat{\pmb{\omega}}_{c}\right)}
\end{equation}

\noindent where $\mathbf{r}$ is the vector from $\mathbf{p}_{a}$ to $\mathbf{p}_{c}$, and $\widehat{\mathbf{r}}$ is its unit vector. Given the actuating angle $\alpha$ defined above, we can represent $\mathbf{r}$ as $\mathbf{r}=d \left( Rot_{y}(\alpha)\left(\begin{matrix}0&0&-1\end{matrix}\right)^{T}\right)$, where $d$ is a pre-set constant to indicate the distance between the capsule and the actuator. Then, $\widehat{\pmb{\omega}}_{a}$ with reference to \{C\} can be written as a function of $\alpha$:

\begin{equation}
\label{F_rotating_actuation_detail}
\widehat{\pmb{\omega}}_{a}=\left(\begin{matrix}\frac{3\sin^{2}\alpha-1}{\sqrt{3\sin^{2}\alpha+1}}&0&\frac{3\cos\alpha\sin\alpha}{\sqrt{3\sin^{2}\alpha+1}}\end{matrix}\right)^{T}
\end{equation}

or

\begin{equation}
\label{F_rotating_actuation_detail_2}
\widehat{\pmb{\omega}}_{a}=Rot_{z}(180^{\circ})Rot_{y}(-\arcsin A)\left(\begin{matrix}1\\0\\0\end{matrix}\right)
\end{equation}
\noindent where $A=\frac{3\cos\alpha\sin\alpha}{\sqrt{3\sin^{2}\alpha+1}}$, $\sqrt{1-A^{2}}=\frac{1-3\sin^{2}\alpha}{\sqrt{3\sin^{2}\alpha+1}}$.

Let $\theta$ represent the angle that $\widehat{\mathbf{m}}_{a}$ rotates about $\widehat{\pmb{\omega}}_{a}$, and we limit the actuating angle $\alpha \in [0^{\circ},35^{\circ}]$ to prevent the actuator from getting too close to the patient, then $\widehat{\mathbf{m}}_{a}$ can be represented as:

\begin{equation}
\label{F_ma}
\begin{aligned}
\widehat{\mathbf{m}}_{a}&= Rot_{z}(180^{\circ})Rot_{y}(-\arcsin A)Rot_{\widehat{\pmb{\omega}}_{a,ini}}(\theta) \widehat{\mathbf{m}}_{a,ini}\\
&=Rot_{z}(180^{\circ})Rot_{y}(-\arcsin A)Rot_{x}(\theta)\left(\begin{matrix}0\\0\\1\end{matrix}\right)\\
&=\left(\begin{matrix} A \cos\theta & \sin\theta & \sqrt{1-A^{2}} \cos\theta \end{matrix}\right)^{T}\\
\end{aligned}
\end{equation}

The rotating magnetic field applied to the capsule $\mathbf{b}_{c}$ can be calculated by the magnetic dipole model \cite{cheng1989field}:

\begin{equation}
\label{F_bc}
\mathbf{b}_{c}(\mathbf{r},\widehat{\mathbf{m}}_{a})=\frac{\mu_{0}\|\mathbf{m}_{a}\|}{4\pi\|\mathbf{r}\|^5}\left(3\mathbf{r}\mathbf{r}^{T}-(\mathbf{r}^{T}\mathbf{r})\mathbf{I}_{3}\right)\widehat{\mathbf{m}}_{a}
\end{equation}

When the capsule is actuated in an environment with no shape restrictions, its magnetic moment $\widehat{\mathbf{m}}_{c}$ can be considered as always parallel to the rotating magnetic field $\mathbf{b}_{c}$. However, this cannot be achieved in the natural lumen pathways of the human body (e.g., the gastrointestinal system) where the moving direction of the capsule is restricted by the inner wall of the tubular environment, then $\widehat{\mathbf{m}}_{c}$ can be represented as:

\begin{equation}
\label{F_mc}
\widehat{\mathbf{m}}_{c}=\Widehat{(\mathbf{b}_{c}-(\mathbf{b}_{c}^{T}\widehat{\pmb{\omega}}_{c}) \widehat{\pmb{\omega}}_{c})}
\end{equation}

\noindent where $(\mathbf{b}_{c}^{T}\widehat{\pmb{\omega}}_{c}) \widehat{\pmb{\omega}}_{c}$ indicates the projection of $\mathbf{b}_{c}$ onto $\widehat{\pmb{\omega}}_{c}$. If $\mathbf{b}_{c}$ is perpendicular to $\widehat{\pmb{\omega}}_{c}$, $\widehat{\mathbf{m}}_{c}$ becomes identical with $\widehat{\mathbf{b}}_{c}$.

As shown in Fig. \ref{Fig_fixedmoving} (b), the magnetic force $\mathbf{f}$ applied to the capsule at $\mathbf{p}_{c}$ with a moving direction of $\widehat{\pmb{\omega}}_{c}$ changes with the actuating angle $\alpha$, and the propulsive force $f_{p}$ is defined as the component of $\mathbf{f}$ in the capsule's moving direction $\widehat{\pmb{\omega}}_{c}$. $\mathbf{f}$ can be calculated by:

\begin{equation}
\label{F_f}
\begin{aligned}
\mathbf{f}(\mathbf{r},\alpha,\theta)&=\frac{3\mu_{0}||\mathbf{m}_{a}|| ||\mathbf{m}_{c}||}{4\pi||\mathbf{r}||^{7}} \\
&\left((\widehat{\mathbf{m}}_{c}\widehat{\mathbf{m}}_{a}^{T})\mathbf{r}(\mathbf{r}^{T}\mathbf{r})
+(\widehat{\mathbf{m}}_{a}\widehat{\mathbf{m}}_{c}^{T})\mathbf{r}(\mathbf{r}^{T}\mathbf{r})\right.\\
&\left.+(\widehat{\mathbf{m}}_{c}^{T}(\mathbf{I}_{3}\mathbf{r}^{T}\mathbf{r}-5\mathbf{r}\mathbf{r}^{T})\widehat{\mathbf{m}}_{a})\mathbf{r}\right)\\
\end{aligned}
\end{equation}

The magnetic force applied to the capsule over one actuator revolution when $\alpha$ is relatively small ($\alpha=10^{\circ}$) is shown in Fig. \ref{Fig_fixedmoving} (c). The propulsive force $f_{p}$ varies little with $\theta$, but the lateral force and the levitation force that are perpendicular to the moving direction show large periodic fluctuations. Therefore, we can approximate the value of $f_{p}$ using its average value $\overline{f}_{p}$ over one actuator revolution by:

\begin{equation}
\label{F_fp}
f_{p}(\mathbf{r}, \alpha) \doteq \overline{f_{p}(\mathbf{r}, \alpha, \theta)}=\frac{1}{2\pi}\int_{0}^{2\pi}\mathbf{f}(\mathbf{r}, \alpha, \theta)^T\widehat{\pmb{\omega}}_{c}\ \mathrm{d}\theta
\end{equation}

Furthermore, Fig. \ref{Fig_fixedmoving} (d) illustrates that the propulsive force $f_{p}$ gradually increases as the actuating angle $\alpha$ becomes larger.

\begin{figure*}[t]
\setlength{\abovecaptionskip}{-0.0cm}
\centering
\includegraphics[scale=1.0,angle=0,width=0.98\textwidth]{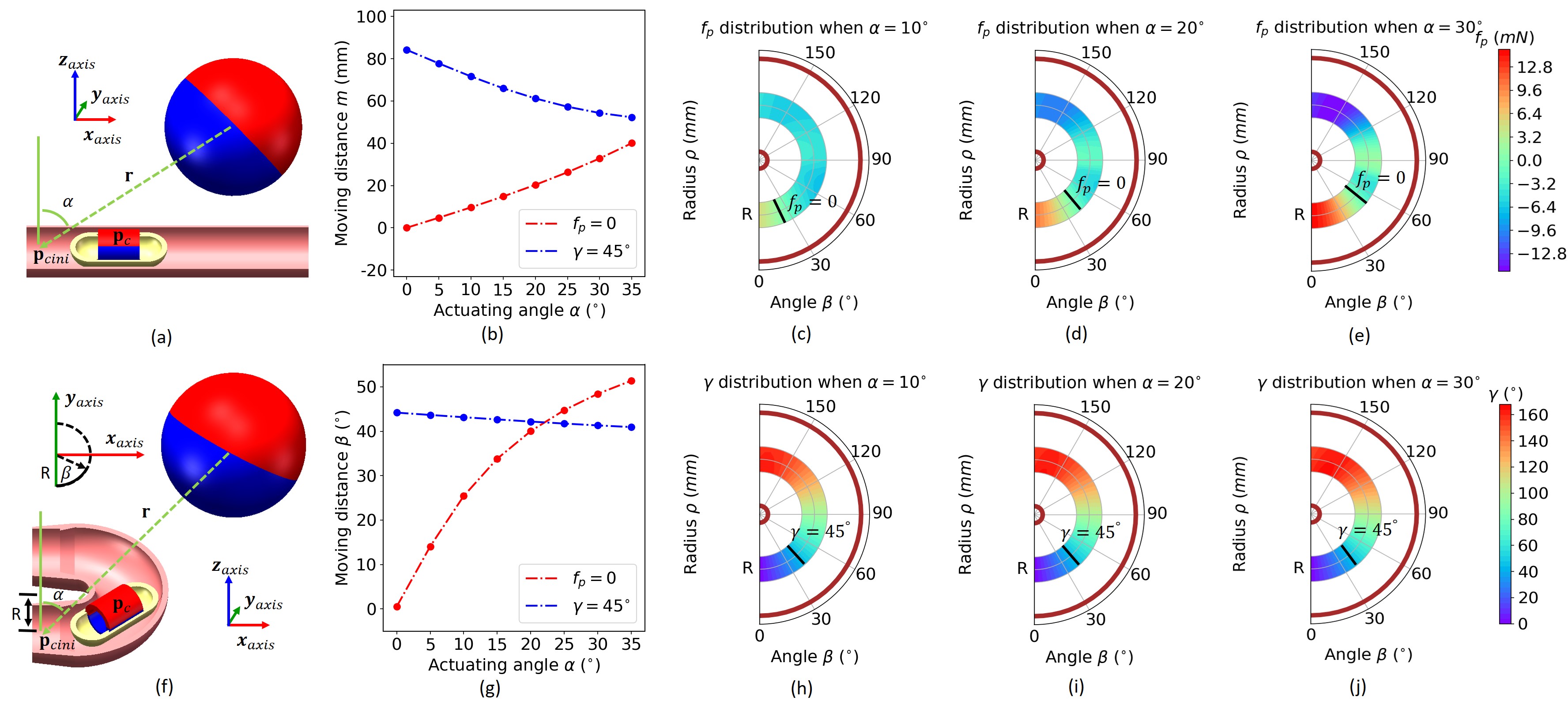}
\caption{(a) and (f) illustrate the propulsion of a capsule in a straight lumen and a $180^{\circ}$ sharp bend, respectively. (b) and (g) show the positions of the zero point of propulsive force ($f_p=0$) and the critical point of valid localization ($\gamma=45^{\circ}$) under different actuating angles ($\alpha$) corresponding to the two environments on the left. (c)-(e) show the distribution of $f_{p}$ in the sharp bend in polar coordinates when $\alpha$ equals $10^{\circ}$, $20^{\circ}$ and $30^{\circ}$, and the black lines indicate the positions where $f_p=0$. (h)-(j) show the distribution of $\gamma$ in the sharp bend in polar coordinates when $\alpha$ equals $10^{\circ}$, $20^{\circ}$ and $30^{\circ}$ , and the black lines indicate the positions where $\gamma=45^\circ$. It can be seen that as $\alpha$ becomes larger, the critical point of valid localization is almost unchanged at $\beta \approx 42^\circ$, while the zero point of propulsive force becomes farther from the initial position, changing from $\beta \approx 25^\circ$ to $\beta \approx 50^\circ$.}
\label{Fig_distribution}
\vspace{-0.4cm}
\end{figure*}

We first consider the situation where the locomotion of the capsule is restricted in a straight lumen (see Fig. \ref{Fig_distribution} (a)).
As shown in Fig. \ref{Fig_fixedmoving} (e), after the pose of the actuator is determined, the capsule will move along its current moving direction $\widehat{\pmb{\omega}}_{c}=\left(\begin{matrix} 1 & 0 & 0 \end{matrix}\right)^{T}$ due to the magnetic force.
Assume the initial position of the capsule is $\mathbf{p}_{cini}=\left(\begin{matrix} 0 & 0 & 0 \end{matrix}\right)^{T}$, and the current position of the capsule is $\mathbf{p}_{c}=\left(\begin{matrix} m & 0 & 0 \end{matrix}\right)^{T}$, then the current relative pose between the capsule and the actuator is $\mathbf{r}_{cur}(m) = \left(\begin{matrix} m - d \sin \alpha & 0 & -d \cos \alpha \end{matrix}\right)^{T}$.
Substituting $\mathbf{r}_{cur}(m)$ and $\widehat{\pmb{\omega}}_{c}$ into (\ref{F_fp}), we can get the relationship between the propulsive force $f_{p}$ and the capsule's moving distance $m$ under different actuating angles, as demonstrated in Fig. \ref{Fig_fixedmoving} (f). Theoretically, the capsule will move to the zero point of the propulsive force ($f_{p}=0$) under the magnetic actuation and oscillate there. Then the localization result at this point (from Section~II-A) will be used to determine the new pose of the actuator. It can be seen that the moving distance of the capsule to this point is farther as $\alpha$ becomes larger. If $\alpha$ is given, the zero point of the propulsive force can be solved by:

\begin{equation}
\label{F_fp_straight}
\begin{aligned}
\mathop{\arg\min}_{m} \quad & |f_{p}(m)|=| \frac{1}{2\pi}\int_{0}^{2\pi}\mathbf{f}(\mathbf{r}_{cur}(m))^T\widehat{\pmb{\omega}}_{c} \ \mathrm{d}\theta | \\
\textrm{subject to} \quad & m\in[0,d\sin{\alpha}]\\
\end{aligned}
\end{equation}

Let $\gamma$ denote the angle difference between the moving direction of the capsule $\widehat{\pmb{\omega}}_{c}$ and the rotation axis $\mathbf{b}_{axis}$ of the rotating magnetic field, which can be calculated as $\mathbf{b}_{axis}=\mathbf{b}_{c}|_{\theta=0^{\circ}}\times\mathbf{b}_{c}|_{\theta=90^{\circ}}$. $\gamma$ can be calculated by:

\begin{equation}
\label{F_gamma}
\gamma=\arccos{\frac{\mathbf{b}_{axis}^{T}\widehat{\pmb{\omega}}_{c}}{||\mathbf{b}_{axis}||}}
\end{equation}

The relationship between $\gamma$ and the moving distance of the capsule $m$ is depicted in Fig. \ref{Fig_fixedmoving} (g). It can be seen that the angle difference $\gamma$ increases with the actuating angle $\alpha$. Through a number of experiments, we found that $\gamma=45^\circ$ is a critical point for effective rotating actuation of the capsule. When $\gamma>45^{\circ}$, the capsule generally cannot rotate synchronously with the actuator, resulting in invalid localization results \cite{xu2020improved}. If $\alpha$ is given, this critical point of valid localization can be solved by:

\begin{equation}
\label{F_gamma_straight}
\begin{aligned}
\mathop{\arg\min}_{m} \quad & |\gamma(m)-45^{\circ}|=|\arccos{\frac{\mathbf{b}_{axis}(m)^{T}\widehat{\pmb{\omega}}_{c}}{||\mathbf{b}_{axis}(m)||}}-45^{\circ}|\\
\textrm{subject to} \quad & m\in[0,d\sin{\alpha}]\\
\end{aligned}
\end{equation}

Fig. \ref{Fig_distribution} (b) shows the positions of the zero point of propulsive force ($f_p=0$) and the critical point of valid localization ($\gamma=45^\circ$) under different actuating angles when the capsule moves in a straight lumen. As we have mentioned, the capsule will move to the position where $f_p=0$ due to the actuating magnetic field. If the localization result at this position is correct ($\gamma<45^\circ$), the next pose of the actuator can be calculated; if the localization result is invalid ($\gamma\geq45^\circ$), the pose of the actuator cannot be updated normally. Therefore, the physical meaning of these two points is actually a trade-off between actuation efficiency and localization accuracy. In a straight lumen, the two curves have no intersections when $\alpha \in [5^{\circ},35^{\circ}]$, which means increasing the actuating angle within this range can improve the actuation efficiency without affecting the localization accuracy.

Next, we consider the case where the capsule moves through a $180^{\circ}$ sharp bend with a radius of $R$, as shown in Fig. \ref{Fig_distribution} (f). Assume the initial position of the capsule is $\mathbf{p}_{cini}=\left(\begin{matrix} 0 & 0 & 0 \end{matrix}\right)^{T}$, we take $\mathbf{p}_{cini}$ as the pole and the negative $y$-axis as the polar axis to construct a polar coordinate system ($\rho$, $\beta$). The current position of the capsule becomes $\mathbf{p}_{c}=\left(\begin{matrix} R\sin\beta & -R\cos\beta & 0 \end{matrix}\right)^{T}$, and the moving direction is $\widehat{\pmb{\omega}}_{c}=\left(\begin{matrix} \cos\beta & \sin\beta & 0 \end{matrix}\right)^{T}$. Then the relative pose between the capsule and the actuator is $\mathbf{r}_{cur}(\beta) = \left(\begin{matrix} R\sin\beta-d\sin\alpha & -R\cos\beta & -d\cos\alpha \end{matrix}\right)^{T}$.
Substituting $\mathbf{r}_{cur}(\beta)$ and $\widehat{\pmb{\omega}}_{c}(\beta)$ into (\ref{F_fp}) and (\ref{F_gamma}), the zero point of propulsive force and the critical point of valid localization when the capsule moves in a sharp bend can be found by solving the following two optimization problems:

\begin{equation}
\label{F_fp_round}
\begin{aligned}
\mathop{\arg\min}_{\beta} \quad & |f_{p}(\beta)|=|\frac{1}{2\pi}\int_{0}^{2\pi}\mathbf{f}(\mathbf{r}_{cur}(\beta))^T\widehat{\pmb{\omega}}_{c}(\beta)\ \mathrm{d}\theta| \\
\textrm{subject to} \quad & \beta\in[0^{\circ},180^{\circ}]\\
\end{aligned}
\end{equation}

\begin{equation}
\label{F_gamma_round}
\begin{aligned}
\mathop{\arg\min}_{\beta} \quad & |\gamma(\beta)-45^{\circ}|=|\arccos{\frac{\mathbf{b}_{axis}(\beta)^{T}\widehat{\pmb{\omega}}_{c}(\beta)}{||\mathbf{b}_{axis}(\beta)||}}-45^{\circ}|\\
\textrm{subject to} \quad & \beta\in[0^{\circ},180^{\circ}]\\
\end{aligned}
\end{equation}

Fig. \ref{Fig_distribution} (g) shows the positions of the zero point of propulsive force ($f_p=0$) and the critical point of valid localization ($\gamma=45^\circ$) under different actuating angles when the capsule moves in the sharp bend. When $\alpha \in [5^{\circ},35^{\circ}]$, the two curves intersect at $\alpha \approx 20^\circ$, which indicates that the actuating angle $\alpha$ should not exceed $20^\circ$ in order to ensure valid localization of the capsule at $f_p=0$. Fig. \ref{Fig_distribution} (c)-(e) and (h)-(j) show the distrubution of $f_{p}$ and $\gamma$ in the $180^\circ$ sharp bend when $\alpha$ equals $10^\circ$,  $20^\circ$ and $30^\circ$ for a clear comparison. When a larger actuating angle $\alpha$ is applied, the critical point of valid localization is almost unchanged at $\beta \approx 42^\circ$, while the zero point of propulsive force is farther away from the initial position, changing from $\beta \approx 25^\circ$ to $\beta \approx 50^\circ$.

It is worth mentioning that if the control frequency of the actuator is high enough, the actuator pose can be updated before the capsule reaches the zero point of propulsive force, and $m$ or $\beta$ will never get close to the point where $\gamma = 45 ^{\circ}$. In this way, the $20^{\circ}$ constraint on $\alpha$ for valid localization can be relaxed. However, since the actuator held by the robotic manipulator in the proposed system cannot realize such a high update frequency, the trade-off between the zero point of propulsive force and the critical point for valid localization ($\gamma = 45 ^{\circ}$) should be taken into consideration.

\subsection{Adaptive adjustment of the actuating angle}

According to the above analysis, we can see that when the capsule is propelled in a straight lumen, increasing the actuating angle can improve the actuation efficiency without affecting the localization accuracy, while in a sharp bend, a smaller actuating angle is required to ensure valid localization results. Therefore, methods for adaptive adjustment of the actuating angle according to the environment should be developed to allow compliant movement of the capsule.

It can be observed that due to the greater resistance caused by the shape limitation, the moving speed of the capsule in the bend is significantly reduced compared with that in the straight lumen. Therefore, we propose to estimate the environment around the capsule (straight or curved lumen) according to its moving speed $v_{c}^{(t)}$, and adaptively adjust the actuating angle to effectively actuate the capsule in both environments.
Let $v_{th}$ denote the velocity threshold used to distinguish between the two environments. When the capsule is moving in a straight lumen with a velocity of $v_{c}^{(t)}>v_{th}$ at time $t$, a greater $\alpha_{H}$ is applied to maximize the actuation efficiency, and when the capsule is moving through a bend with a velocity of $v_{c}^{(t)} \leq v_{th}$, a smaller $\alpha_{L}$ is used to ensure reliable localization.
The adaptive actuation method is summarized by:

\begin{equation}
\label{F_adaptive_force}
\alpha^{(t)}=\left\{
\begin{aligned}
\alpha_{H} &,\  v_{c}^{(t)} > v_{th}\\
\alpha_{L} &,\  v_{c}^{(t)} \leq v_{th}\\
\end{aligned}\right.
\end{equation}

In practice, since the pose of the actuator is changing in real time, the estimation of the instantaneous velocity of the capsule will fluctuate due to the localization noise. Therefore, we calculate the \textit{Simple Moving Average (SMA)} of the capsule's moving speed with a period of $3 \Delta T$ to approximate the capsule's moving speed at time $t$ by:

\vspace{-0.2cm}
\begin{equation}
\label{F_moving_average}
\begin{aligned}
v_{c}^{(t)} \leftarrow SMA_3(v_{c}^{(t)})&=\frac{1}{3}\sum_{i=0}^2 v_{c}^{(t-i \Delta T)}\\
&=\frac{\{v_{c}^{(t-2 \Delta T)}+v_{c}^{(t-\Delta T)}+v_{c}^{(t)}\}}{3}
\end{aligned}
\end{equation}

\noindent here the moving speed of the capsule at time $t-i \Delta T$ is estimated using $v_{c}^{(t-i \Delta T)} = \frac{\|\mathbf{p}_{c}^{(t-i \Delta T)}-\mathbf{p}_{c}^{(t-(i+1) \Delta T)}\|}{\Delta T}$, $i \in \{0,1,2\}$.

\subsection{Adaptive actuation algorithm}

The adaptive actuation algorithm is summarized in Algorithm \ref{Alg_act} and illustrated in Fig. \ref{Fig_workflow} (b).

\begin{algorithm} 
\caption{Adaptive Actuation Algorithm}
\label{Alg_act}
\KwIn{current $\widehat{\pmb{\omega}}_{c}^{(t)}$ and $4$ sampled positions with time step $\Delta T$, $\{\mathbf{p}_{c}^{(t-3 \Delta T)}, \mathbf{p}_{c}^{(t-2 \Delta T)}, \mathbf{p}_{c}^{(t-\Delta T)}, \mathbf{p}_{c}^{(t)}\}$}
\KwOut{updated actuator's pose $\mathbf{p}_{a}^{(t)}$, $\widehat{\pmb{\omega}}_{a}^{(t)}$}
Receive $4$ sampled positions of the capsule with time step $\Delta T$  $\{\mathbf{p}_{c}^{(t-3 \Delta T)}, \mathbf{p}_{c}^{(t-2 \Delta T)}, \mathbf{p}_{c}^{(t-\Delta T)}, \mathbf{p}_{c}^{(t)}\}$  from the adaptive localization pipeline\;
\For{$i=0,1,2$}
{Estimate the velocity at time $t-i \Delta T$ using $v_{c}^{(t-i \Delta T)} = \frac{\|\mathbf{p}_{c}^{(t-i \Delta T)}-\mathbf{p}_{c}^{(t-(i+1) \Delta T)}\|}{\Delta T}$\;}
Update the current velocity of the capsule $v_{c}^{(t)}$ using simple moving average by (\ref{F_moving_average})\;
\If{$v_{c}^{(t)} > v_{th}$}
{$\alpha^{(t)}=\alpha_{H}$\;}
\Else{
$\alpha^{(t)}=\alpha_{L}$\;
}
Calculate $\mathbf{r}^{(t)}=d \left( Rot_{y}(\alpha^{(t)})\left(\begin{matrix}0&0&-1\end{matrix}\right)^{T}\right)$ \;
Update the position of the actuator $\mathbf{p}_{a}^{(t)}=\mathbf{p}_{c}^{(t)}-\mathbf{r}^{(t)}$\;
Update the rotation axis of the actuator $\widehat{\pmb{\omega}}_{a}^{(t)}$ by (\ref{F_rotating_actuation_detail})\;
\Return {updated actuator's pose} $\mathbf{p}_{a}^{(t)}$, $\widehat{\pmb{\omega}}_{a}^{(t)}$\;
\end{algorithm}

\section{Experiments and Results}

\subsection{Implementation Details}

\begin{figure}[b]
\centering
\includegraphics[scale=1.0,angle=0,width=0.42\textwidth]{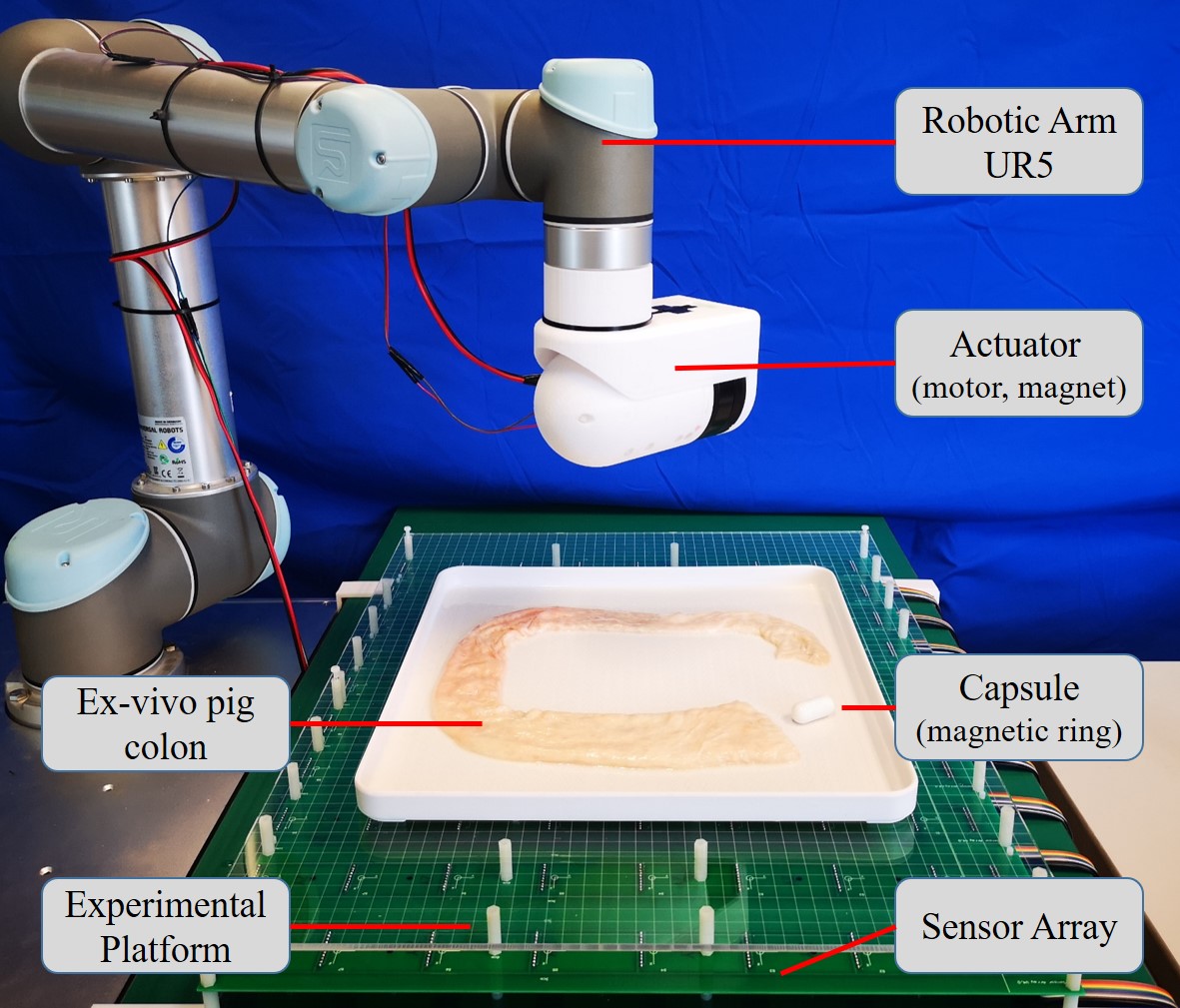}
\caption{Setup of our proposed SMAL system for experiments in an ex-vivo pig colon. The actuator comprised of a motor and a spherical magnet is mounted at the end-effector of a UR5 robotic arm. The capsule with a embedded magnetic ring is actuated in the pig colon on a horizontal experimental platform over a large sensor array.}
\label{Fig_implementation_system}
\end{figure}

Fig. \ref{Fig_implementation_system} shows the setup of our proposed SMAL system for experiments in an ex-vivo pig colon. A 6-DoF serial robotic manipulator (5-kg payload, UR5, Universal Robots) is used to control the movement of the actuator mounted at its end-effector, which is comprised of a motor (RMD-L-90, GYEMS) and a spherical permanent magnet (diameter $50mm$, NdFeB, N42 grade). The capsule (diameter $16mm$, length $35mm$) comprises a 3D-printed shell (Polylactic Acid, UP300 3D printer, Tiertime) and a permanent magnetic ring (outer diameter $12.8mm$, inner diameter $9mm$, and length $15mm$, NdFeB, N38SH grade). The large external sensor array includes $80$ three-axis magnetic sensors (MPU9250, InvenSense) arranged in an $8\times10$ grid with a spacing of $6cm$ to cover the entire abdominal region of the patient. The output frequency of each sensor is $100Hz$. We use $10$ USB-I2C adaptors (Ginkgo USB-I2C, Viewtool), a USB-CAN adaptor (Ginkgo USB-CAN, Viewtool) and a network cable for data transmission. The SMAL algorithm is implemented with Python and runs on a desktop (Intel i7-7820X, 32GB RAM, Win10).

\subsection{Evaluation of Adaptive localization}

\begin{figure*}[t]
\setlength{\abovecaptionskip}{-0.0cm}  
\centering
\includegraphics[scale=1.0,angle=0,width=0.99\textwidth]{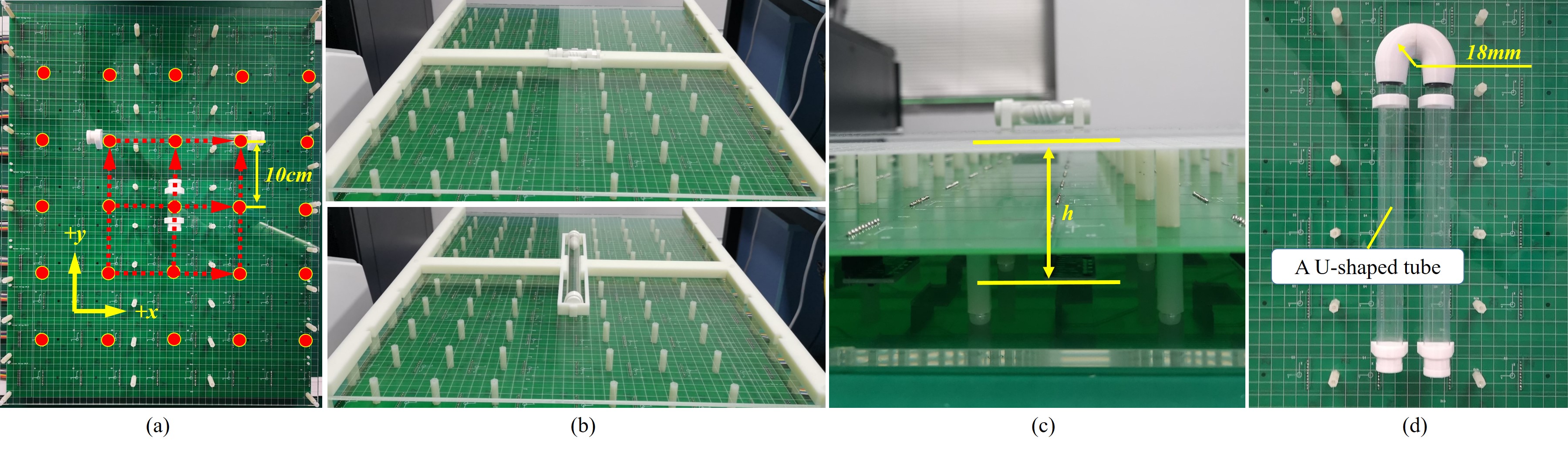}
\caption{(a) For evaluation of the adaptive localization algorithm, the ``fixed point" experiments are conducted by actuating the capsule in a short plastic tube which is placed at $25$ red points with an interval of $10cm$, and the ``moving" experiments are conducted by actuating the capsule in a long plastic tube placed along $6$ red dotted lines in the $+x$ and $+y$ directions. (b) The tube and the platform are rigidly connected using 3D-printed connectors during the ``fixed point" and ``moving" experiments. (c) Localization experiments at different heights.(d) A U-shaped tube with a radius of $18mm$ is used to evaluate the proposed adaptive actuation method.}
\label{Fig_expfixedmoving}
\end{figure*}

\begin{table}[b] \footnotesize
\centering
\caption{Localization accuracy with different layouts}
\begin{tabular}{|p{0.8cm}<{\centering}|p{2.2cm}<{\centering}|p{2.2cm}<{\centering}|p{1.9cm}<{\centering}|}
\hline
\multirow{4}{*}{\tabincell{c}{Sensor\\Array\\Layout}} & \multicolumn{2}{c|}{\multirow{2}{*}{\tabincell{c}{Localization Error \\ (Position / Orientation)}}} & \multirow{4}{*}{\tabincell{c}{Processing\\Time ($ms$)}} \\
{} & \multicolumn{2}{c|}{} & {} \\
\cline{2-3}
{} & \multirow{2}{*}{\tabincell{c}{``Fixed point"\\experiments}} & \multirow{2}{*}{\tabincell{c}{``Moving"\\experiments}} & {} \\
{} & {} & {} & {} \\
\hline
\multirow{2}{*}{Entire} & \multirow{2}{*}{\tabincell{c}{$268.2\pm25.6$ $mm$  /\\ $42.7\pm31.2$ $^{\circ}$}} & \multirow{2}{*}{\tabincell{c}{$287.1\pm27.5$ $mm$ /\\ $46.2\pm42.1$ $^{\circ}$}} & \multirow{2}{*}{$144.4\pm11.6$} \\
{} & {} & {} & {} \\
\hline
\multirow{2}{*}{\cite{xu2020improved}} & \multirow{2}{*}{\tabincell{c}{$3.3\pm1.3$ $mm$ /\\$6.7\pm1.1$ $^{\circ}$}} & \multirow{2}{*}{\tabincell{c}{$4.3\pm1.7$ $mm$ /\\$9.3\pm1.0$ $^{\circ}$}} & \multirow{2}{*}{$32.1\pm2.8$} \\
{} & {} & {} & {} \\
\hline
\multirow{2}{*}{(a)} & \multirow{2}{*}{\tabincell{c}{$4.8\pm2.1$ $mm$ /\\$4.6\pm1.1$ $^{\circ}$}} & \multirow{2}{*}{\tabincell{c}{$5.0\pm2.1$ $mm$ /\\$5.8\pm2.8$ $^{\circ}$}} & \multirow{2}{*}{$19.7\pm3.1$} \\
{} & {} & {} & {} \\
\hline
\multirow{2}{*}{(b)} & \multirow{2}{*}{\tabincell{c}{$6.5\pm2.8$ $mm$ /\\$6.0\pm2.9$ $^{\circ}$}} & \multirow{2}{*}{\tabincell{c}{$6.9\pm2.4$ $mm$ /\\$7.8\pm2.1$ $^{\circ}$}} & \multirow{2}{*}{$19.6\pm2.7$} \\
{} & {} & {} & {} \\
\hline
\multirow{2}{*}{(c)} & \multirow{2}{*}{\tabincell{c}{$\mathbf{4.2\pm1.8}$ $mm$ /\\$\mathbf{4.2\pm1.2}$ $^{\circ}$}} & \multirow{2}{*}{\tabincell{c}{$\mathbf{4.4\pm2.1}$ $mm$ /\\$\mathbf{6.6\pm2.3}$ $^{\circ}$}} & \multirow{2}{*}{$\mathbf{19.5\pm2.2}$} \\
{} & {} & {} & {} \\
\hline
\multirow{2}{*}{(d)} & \multirow{2}{*}{\tabincell{c}{$4.4\pm1.8$ $mm$ /\\$4.5\pm1.0$ $^{\circ}$}} & \multirow{2}{*}{\tabincell{c}{$4.8\pm1.7$ $mm$ /\\$6.6\pm2.0$ $^{\circ}$}} & \multirow{2}{*}{$21.1\pm2.6$} \\
{} & {} & {} & {} \\
\hline
\multirow{2}{*}{(e)} & \multirow{2}{*}{\tabincell{c}{$5.9\pm2.7$ $mm$ /\\$7.1\pm3.2$ $^{\circ}$}} & \multirow{2}{*}{\tabincell{c}{$6.5\pm3.1$ $mm$ /\\$9.0\pm3.3$ $^{\circ}$}} & \multirow{2}{*}{$21.6\pm3.3$} \\
{} & {} & {} & {} \\
\hline
\multirow{2}{*}{(f)} & \multirow{2}{*}{\tabincell{c}{$5.9\pm2.4$ $mm$ /\\$6.3\pm2.5$ $^{\circ}$}} & \multirow{2}{*}{\tabincell{c}{$6.6\pm3.8$ $mm$ /\\$8.8\pm3.7$ $^{\circ}$}} & \multirow{2}{*}{$21.7\pm3.5$} \\
{} & {} & {} & {} \\
\hline
\end{tabular}
\label{T_expfixedmoving}
\end{table}

The localization accuracy of the system is evaluated in two sets of experiments, as shown in Fig. \ref{Fig_expfixedmoving} (a). First, the capsule is actuated in a short straight PVC tube (length $35.3mm$, inner diameter $18mm$), whose size is slightly larger than the capsule to make the position and heading direction of the capsule remain unchanged during actuation. These ``fixed point" experiments are conducted at $25$ points with an interval of $10cm$. 
To evaluate the localization accuracy of a capsule during movement, the capsule is then actuated in a long straight PVC tube (length $200mm$, inner diameter $18mm$) placed along $6$ lines in the $+x$ and $+y$ directions, so its movement is limited to translation along and rotation around the axis of the tube. This set of experiments are referred to as the ``moving" experiments.
In order to quantitatively evaluate the localization method, the 3D geometry of the experimental platform and the PVC tubes are specially designed and manufactured, and the relative pose between the tube and the platform is manually determined. During the experiments, the tube and the platform are firmly and rigidly connected using 3D-printed connectors, as shown in Fig. \ref{Fig_expfixedmoving} (b). The rigid connection not only prevents the tube from sliding on the platform during the experiment, but is also used to provide the ground truth of the capsule position and heading direction. The errors of manufacturing and 3D printing are less than $0.1mm$.

The criteria to evaluate the sensor layouts in real-world applications should include both the localization accuracy and the data processing time to facilitate accurate localization with high update frequency during the active propulsion of the capsule. Table \ref{T_expfixedmoving} illustrates the localization accuracy and processing time of different sensor array layouts in the real-world experiments, including the entire array ($8\times10$ layout), the $4\times4$ layout in \cite{xu2020improved}, and our six selected layout candidates (a)-(f) as shown in Fig. \ref{Fig_layouts_candidates}. In terms of localization accuracy, layout (c) which is selected as the optimal layout in Section~III-A achieves the highest average localization accuracy of $4.3 mm / 5.4^{\circ}$ in the real-world experiments, which is consistent with our simulation. Meanwhile, as layout (c) contains the fewest sensors in the sub-array (only $8$ sensors), it takes the shortest time to read and process the sensor data ($< 20ms$) during each localization update. Since we want to enhance the real-time capabilities of the system, the solution that uses fewer sensors and provides higher localization accuracy is preferred, which can accelerate data processing and simplify the hardware implementation.

\begin{table}[t] \footnotesize
\centering
\caption{Localization accuracy at different heights}
\begin{tabular}{|p{1.2cm}<{\centering}|p{3.0cm}<{\centering}|p{3.0cm}<{\centering}|}
\hline
\multirow{4}{*}{\tabincell{c}{Height $h$}} & \multicolumn{2}{c|}{\multirow{2}{*}{\tabincell{c}{Localization Error (Position / Orientation)}}} \\
{} & \multicolumn{2}{c|}{} \\
\cline{2-3}
{} & \multirow{2}{*}{\tabincell{c}{``Fixed point" experiments}} & \multirow{2}{*}{\tabincell{c}{``Moving" experiments}} \\
{} & {} & {} \\
\hline
\multirow{2}{*}{$44$ $mm$} & \multirow{2}{*}{\tabincell{c}{$4.2\pm1.8$ $mm$ /\\$4.2\pm1.2$ $^{\circ}$}} & \multirow{2}{*}{\tabincell{c}{$4.4\pm2.1$ $mm$ /\\$6.6\pm2.3$ $^{\circ}$}} \\
{} & {} & {} \\
\hline
\multirow{2}{*}{$64$ $mm$} & \multirow{2}{*}{\tabincell{c}{$4.3\pm1.6$ $mm$ /\\$4.1\pm1.5$ $^{\circ}$}} & \multirow{2}{*}{\tabincell{c}{$4.5\pm1.1$ $mm$ /\\$6.4\pm3.1$ $^{\circ}$}} \\
{} & {} & {} \\
\hline
\multirow{2}{*}{$84$ $mm$} & \multirow{2}{*}{\tabincell{c}{$4.3\pm2.1$ $mm$ /\\$4.5\pm2.1$ $^{\circ}$}} & \multirow{2}{*}{\tabincell{c}{$4.5\pm2.2$ $mm$ /\\$6.6\pm2.5$ $^{\circ}$}} \\
{} & {} & {} \\
\hline
\multirow{2}{*}{$104$ $mm$} & \multirow{2}{*}{\tabincell{c}{$5.6\pm2.7$ $mm$ /\\$6.5\pm3.2$ $^{\circ}$}} & \multirow{2}{*}{\tabincell{c}{$5.9\pm2.8$ $mm$ /\\$7.2\pm3.2$ $^{\circ}$}} \\
{} & {} & {} \\
\hline
\multirow{2}{*}{$124$ $mm$} & \multirow{2}{*}{\tabincell{c}{$6.1\pm2.3$ $mm$ /\\$7.0\pm2.9$ $^{\circ}$}} & \multirow{2}{*}{\tabincell{c}{$6.3\pm3.2$ $mm$ /\\$9.2\pm3.8$ $^{\circ}$}} \\
{} & {} & {} \\
\hline
\multirow{2}{*}{$144$ $mm$} & \multirow{2}{*}{\tabincell{c}{$11.5\pm5.6$ $mm$ /\\$75.3\pm36.2$ $^{\circ}$}} & \multirow{2}{*}{\tabincell{c}{$15.2\pm6.8$ $mm$ /\\$78.5\pm42.5$ $^{\circ}$}} \\
{} & {} & {} \\
\hline
\end{tabular}
\label{T_differentheight}
\end{table}

We further conduct six sets of localization experiments using the optimal sensor sub-array layout to assess the performance of our method at different heights (from $44mm$ to $144mm$), as shown in Fig. \ref{Fig_expfixedmoving} (c) and Table \ref{T_differentheight}. It can be seen from the results that the localization errors can be maintained on the order of millimeter when the capsule height is chosen below $124mm$. This is because when the capsule is placed farther from the sensor array, it is more difficult to obtain accurate measurement of the magnetic field of the capsule, which will result in a decrease in the localization accuracy. Therefore, in the current version of the system, the capsule should be placed no more than $12cm$ above the sensor array. In order to further expand the workspace along the z-direction, one can use magnetic sensors with better performance or increase the magnetic moment of the embedded magnet in the capsule. In addition, since the sensor array in the current system is designed in a 2D layout, future methods may investigate a 3D sensor array (e.g., in \cite{hu2010new}) to reduce the restrictions on the height of the capsule.

\subsection{Evaluation of adaptive actuation}

\begin{figure}[b]
\setlength{\abovecaptionskip}{-0.1cm}  
\centering
\includegraphics[scale=1.0,angle=0,width=0.48\textwidth]{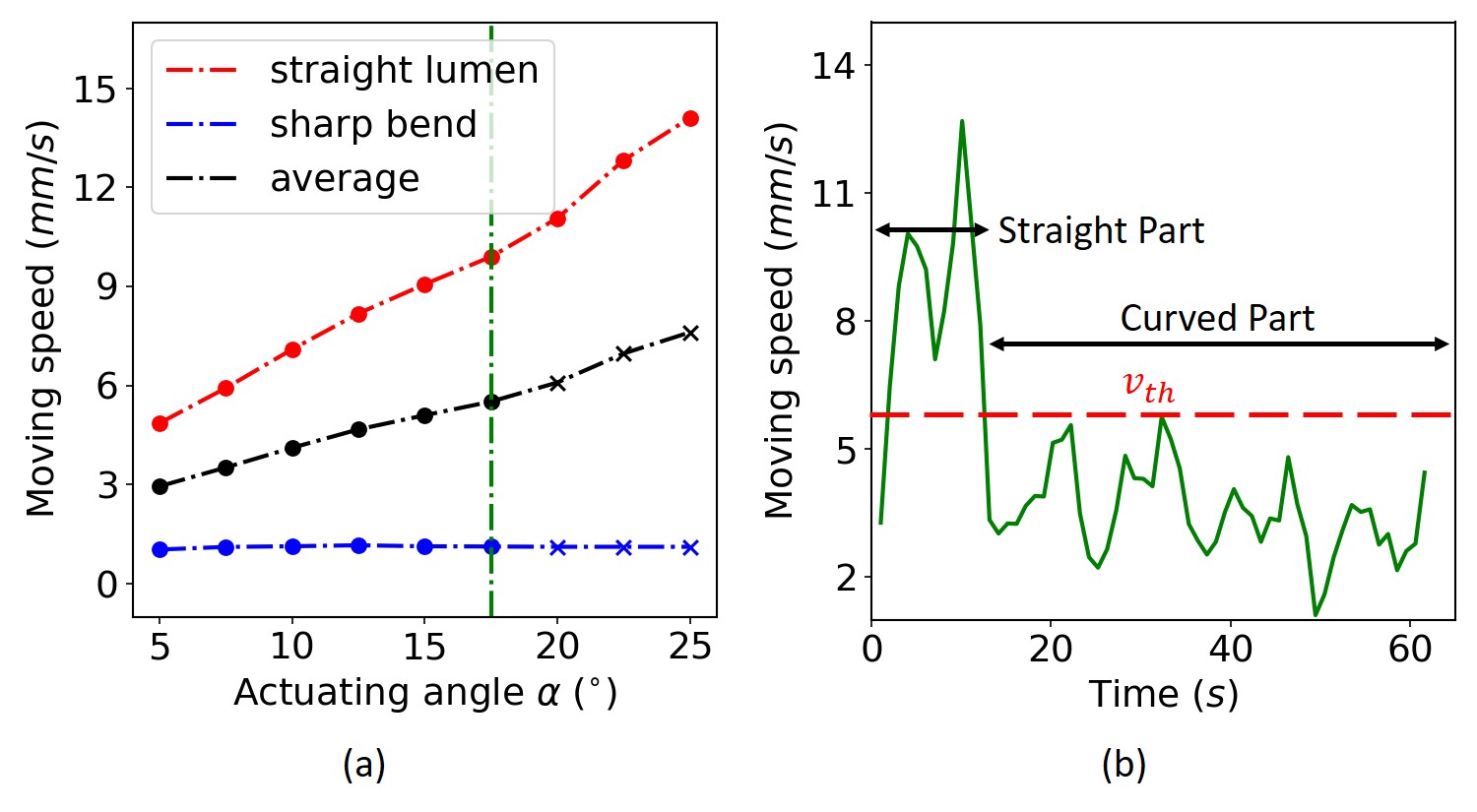}
\caption{(a) The capsule is separately actuated in the straight part and the bend of a U-shaped tube under different actuating angles changing from $5^{\circ}$ to $25^{\circ}$. $10$ experiments are conducted for each $\alpha$. The average moving speed of the capsule in the straight part and the bend are marked as red and blue points, respectively. The black dots indicate their average values. The capsule fails to pass through the bend when $\alpha$ is greater than $17.5^{\circ}$ (green dashed lines), and its moving speed is marked as a cross. (b) shows the capsule's moving speed in the U-shaped tube when $\alpha=15^\circ$. Therefore, a threshold $v_{th}=5.7$ is chosen to distinguish between the straight part and the curved part.}
\label{Fig_expalphastudy}
\end{figure}

As shown in Fig. \ref{Fig_expfixedmoving} (d), the capsule is actuated in a U-shaped tube with an inner diameter of $18mm$, and the radius of the $180^{\circ}$ bend is $18mm$.
As shown in Fig. \ref{Fig_expalphastudy} (a), as the actuating angle $\alpha$ changes from $5^{\circ}$ to $25^{\circ}$, the average moving speed of the capsule in the straight part of the tube increases from $4.9mm/s$ to $14.1mm/s$, while in the curved part, the capsule keeps moving at a low speed of about $1mm/s$. The capsule fails to pass through the sharp bend when $\alpha$ is greater than $17.5^{\circ}$, which is very close to the theoretical value ($\approx 20^\circ$) as shown in Fig. \ref{Fig_distribution} (g).

\begin{figure}[t]
\setlength{\abovecaptionskip}{-0.1cm}  
\centering
\includegraphics[scale=1.0,angle=0,width=0.49\textwidth]{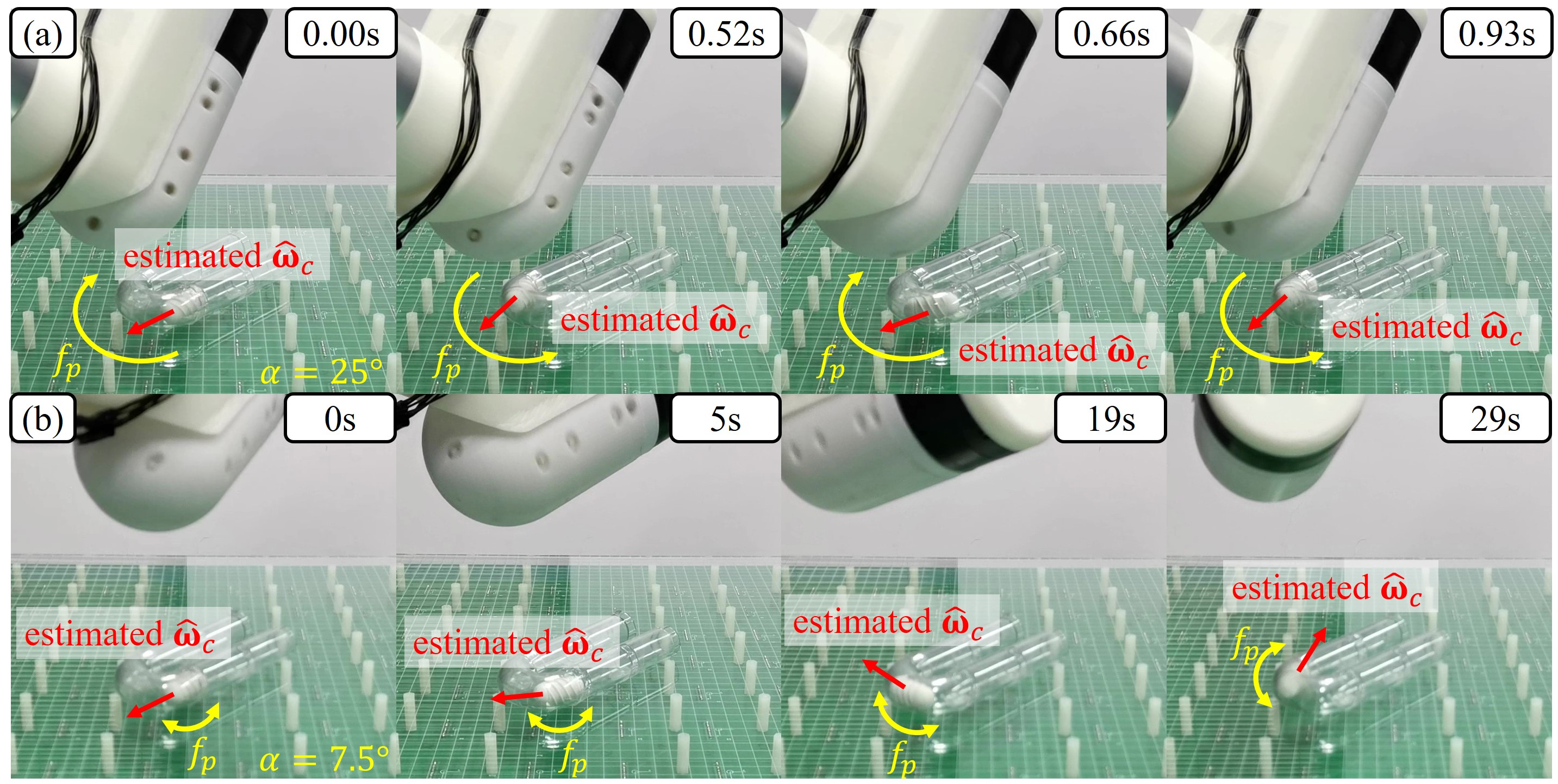}
\caption{The capsule is actuated through a U-shaped tube when (a) $\alpha=25^{\circ}$ and (b) $\alpha=7.5^{\circ}$. The yellow arrow indicates the magnitude and direction of the propulsive force, and the red arrow indicates the estimated heading direction of the capsule.}
\label{Fig_failure}
\end{figure}

In order to analyse the failure mode under a large actuating angle, we conduct comparative experiments in a U-shaped tube when $\alpha$ is set to $25^\circ$ and $7.5^\circ$, respectively. As shown in Fig. \ref{Fig_failure} (a), when $\alpha$ is set to $25^\circ$, the large propulsive force will cause the capsule to suddenly slide through the $180^\circ$ bend and result in an abrupt change in the pose of the capsule. Since $\gamma > 45^\circ$, the capsule cannot rotate synchronously with the actuator, and the estimated heading direction of the capsule remains almost unchanged. Therefore, the updated actuator's pose is kept almost unchanged. Then, the capsule is immediately pulled back to the original position under a large negative propulsive force as shown in Fig. \ref{Fig_distribution} (e), and again, slides through the bend under the large propulsive force. In this way, the capsule keeps reciprocating between these two positions and fails to pass the bend. In contrast, as shown in Fig. \ref{Fig_failure} (b), when a small actuating angle is applied ($\alpha=7.5^\circ$), the capsule moves at a lower speed under a smaller propulsive force (as shown in Fig. \ref{Fig_distribution} (c)) in the $180^\circ$ bend and can rotate normally under magnetic actuation. Since $\gamma < 45^\circ$, the capsule can be located correctly, and the actuator's pose can be updated correctly to stably propel the capsule through the bend.

Therefore, $\alpha_{L}$ and $\alpha_{H}$ in (\ref{F_adaptive_force}) should be selected to be smaller than $17.5^\circ$ to ensure that the capsule can be located correctly when it enters the curved part of the tube. Meanwhile, $\alpha_{H}$ should be as large as possible to improve the actuation efficiency in the straight part. Based on the above considerations, we choose $\alpha_{H}=15^\circ$.

In order to correctly identify the straight and curved parts of the U-shaped tube and then use (\ref{F_adaptive_force}) to adaptively change the actuating angle, an appropriate velocity threshold $v_{th}$ needs to be determined. When $\alpha=15^\circ$, the capsule's moving speed (\textit{Simple Moving Average}) in the U-shaped tube is shown in Fig. \ref{Fig_expalphastudy} (b). It can be seen that there is a significant difference in the moving speed of the capsule in the straight part and the curved part. Therefore, we choose $v_{th}=5.7$ as the threshold to distinguish the two environments.

\begin{figure}[t]
\setlength{\abovecaptionskip}{-0.0cm}  
\centering
\includegraphics[scale=1.0,angle=0,width=0.46\textwidth]{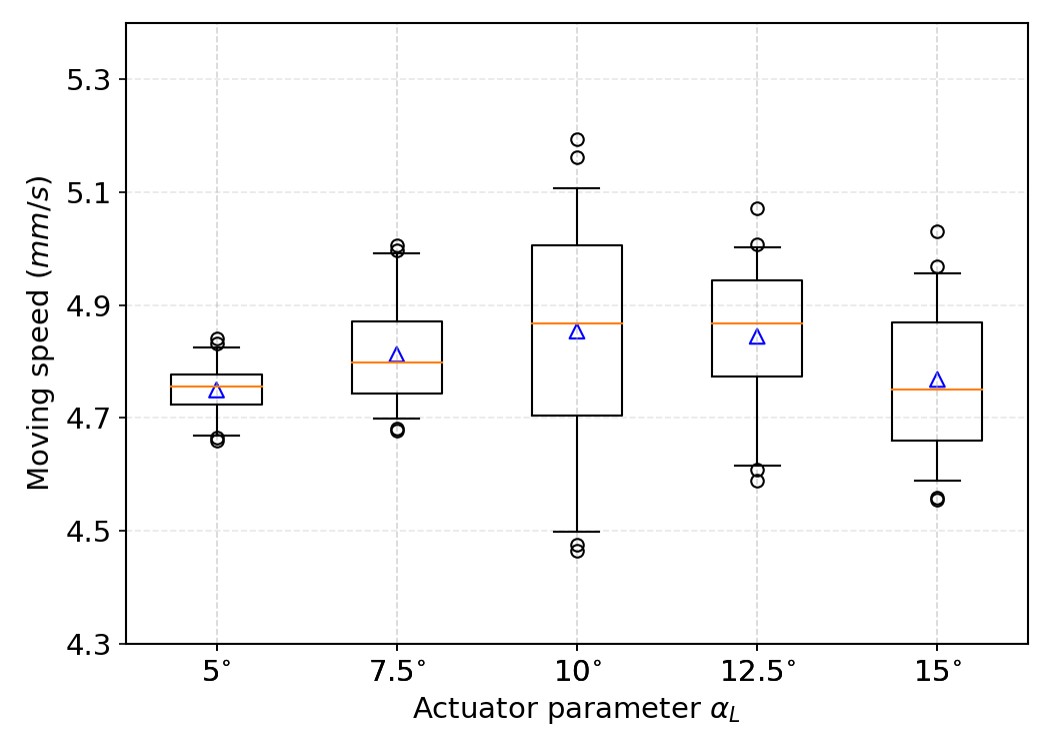}
\caption{Moving speed of the capsule in a U-shaped tube when $\alpha_{H}=15^\circ$, $v_{th}=6$, and $\alpha_{L}$ is chosen as $5^\circ$, $7.5^\circ$, $10^\circ$, $12.5^\circ$ and $15^\circ$. $30$ experiments are conducted for each parameter set. When $\alpha_{L} \leq 10^\circ$, the mean value of the moving speed gradually increases, but the variance between experiments also becomes larger. When $\alpha_{L}>10^\circ$, the mean value of the moving speed decreases. This is because as the actuating angle increases, the zero point of propulsive force and the critical point of valid localization in the curved part become closer, resulting in a decrease in the stability of actuation.}
\label{Fig_expalphastudy2}
\end{figure}

When $\alpha_{H}=15^\circ$ and $v_{th}=5.7$, we compare the average moving speed of the capsule in the U-shaped tube when $\alpha_{L}$ is chosen as $5^\circ$, $7.5^\circ$, $10^\circ$, $12.5^\circ$ and $15^\circ$. As shown in Fig. \ref{Fig_expalphastudy2}, when $\alpha_{L} \leq 10^\circ$, the mean value of the moving speed gradually increases, but the variance between experiments also becomes larger. When $\alpha_{L}>10^\circ$, the mean value of the moving speed decreases and the variance remains large. This is because as the actuating angle increases, the zero point of propulsive force and the critical point of valid localization in the curved part become closer, resulting in a decrease in the stability of actuation. The results are consistent with the theoretical analysis in Section~IV-A. Therefore, $\alpha_L=7.5^\circ$ is a good informed choice considering both the mean and variance. Meanwhile, in order to allow the algorithm to identify when the capsule has left a bend and entered a straight section to switch $\alpha$ from $\alpha_{L}$ to $\alpha_{H}$, the moving speed of the capsule in the straight section under $\alpha_{L}$ should be larger than $v_{th}$. When $\alpha_{L}=7.5^\circ$, this requirement is satisfied as the speed of the capsule in the straight lumen is $6.2mm/s$, as shown in Fig. \ref{Fig_expalphastudy} (a). Therefore, we finally choose $\alpha_{L}=7.5^\circ$.

\subsection{Overall closed-loop SMAL performance in complex environments}

\begin{figure*}[t]
\centering
\includegraphics[scale=1.0,angle=0,width=0.98\textwidth]{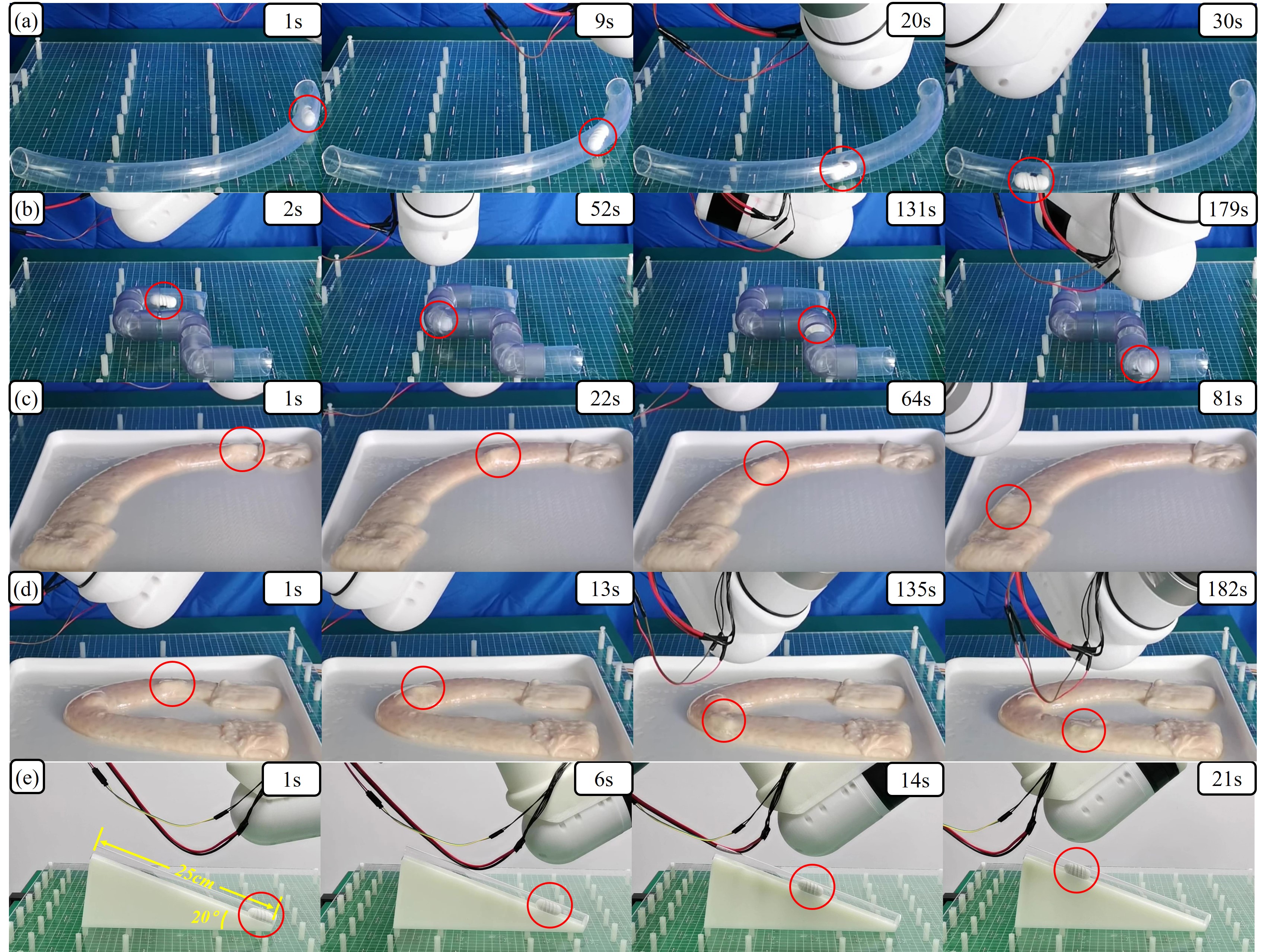}
\caption{Three demonstrations are conducted in (a) a long curved PVC tube, (b) a PVC tube with continuous bends, (c) a curved ex-vivo pig colon, (d) a U-shaped ex-vivo pig colon, and (e) a straight PVC tube  placed on a $20^\circ$ slope. The capsule location in each image is highlighted using a red circle.}
\label{Fig_expdemo}
\end{figure*}

In order to further demonstrate the effectiveness of the overall approach and see the closed-loop SMAL performance of the system in more complex environments, we conduct additional experiments in a long curved PVC tube, a PVC tube with continuous bends, a curved ex-vivo pig colon and a U-shaped ex-vivo pig colon. We apply the same actuating angles $\alpha_{H}=15^{\circ}$ and $\alpha_{L}=7.5^{\circ}$ in all these experiments.
In Fig. \ref{Fig_expdemo} (a), the capsule is propelled through a long curved PVC tube (length $412 mm$, inner diameter $25 mm$). It takes $41 s$ and the average moving speed of the capsule is $10.05 mm/s$.
In Fig. \ref{Fig_expdemo} (b), the capsule moves through a PVC tube with continuous bends (length $397 mm$, inner diameter $25 mm$). The average moving speed of the capsule is $2.34 mm/s$, which is significantly reduced by $76.72$\% compared with (a) due to the increase in the bending angle of the tube and the more complicated shape.
In Fig. \ref{Fig_expdemo} (c), the capsule is actuated in a curved ex-vivo pig colon (length $250 mm$). The average moving speed of the capsule is $3.13 mm/s$, which declines by $68.86$\% compared with (a) since the resistance of the pig colon is significantly increased compared with the PVC tube.
In Fig. \ref{Fig_expdemo} (d), the capsule is actuated in a U-shaped ex-vivo pig colon (length $250 mm$). The average moving speed of the capsule is $1.44 mm/s$, which is reduced by $54.00$\% compared with (c) as the increased bending angle further reduces the speed of the capsule.
In Fig. \ref{Fig_expdemo} (e), the capsule is propelled in a straight PVC tube (length $250 mm$, inner diameter $18mm$) placed on a $20^\circ$ slope. The average moving speed of the capsule is $9.04 mm/s$, which is slightly slower than the speed in the long curved tube placed on the horizontal plane. This result qualitatively demonstrates that the proposed system can realize spatial motion of the capsule in the z-component of the workspace.
It can be seen that although the shape and material of different environments will affect the moving speed of the capsule, our method can successfully complete the simultaneous actuation and localization tasks and smoothly propel the capsule through the complex tubular environments, which demonstrates the robustness of our proposed system. 

Assuming that the moving speed of the capsule in the human intestine is similar with that in the pig colon, it can be roughly estimated that it would take about $69min$ to complete the active WCE examination in the small intestine of approximate $6m$ \cite{hounnou2002anatomical} with our method, which is much shorter than the examination time using the current passive WCE (about $247.2min$) \cite{liao2010fields}\cite{ge2007clinical}. Therefore, the SMAL system proposed in this work will hopefully shorten the examination time of WCE, thereby improving the clinical acceptance of this non-invasive and painless screening technique.

\begin{table*}[t] \footnotesize
\centering
\caption{Overall Performance Comparison of State-of-the-art Permanent Magnet-based SMAL Systems}
\begin{tabular}{|p{2.7cm}<{\centering}|p{1.8cm}<{\centering}|p{2.3cm}<{\centering}|p{2.3cm}<{\centering}|p{1.9cm}<{\centering}|p{1.9cm}<{\centering}|p{2.1cm}<{\centering}|}
\hline
Studies & Song et al. \cite{song2016real} & Taddese et al. \cite{taddese2018enhanced} & Popek et al. \cite{popek2017first} & Xu et al. \cite{xu2020novelsystem} & Xu et al. \cite{xu2020improved} & Ours \\
\hline
Actuator & \tabincell{c}{Dragging\\permanent\\magnet} & \tabincell{c}{Dragging\\permanent\\magnet with\\electromagnetic\\coils} & \tabincell{c}{Rotating\\permanent\\magnet} & \tabincell{c}{Rotating\\permanent\\magnet} & \tabincell{c}{Rotating\\permanent\\magnet} & \tabincell{c}{\textbf{Rotating}\\ \textbf{permanent}\\ \textbf{magnet}} \\
\hline
\tabincell{c}{Inside the capsule} & {$1$ magnet}  & \tabincell{c}{$1$ cube magnet\\with $6$ sensors} & \tabincell{c}{$1$ cube magnet\\with $6$ sensors} & \tabincell{c}{$2$ magnetic\\rings} & {$1$ magnetic ring} & {$\mathbf{1}$ \textbf{magnetic ring}} \\
\hline
\tabincell{c}{Sensor location} & External & Internal & Internal & External & External & \textbf{External} \\
\hline
\tabincell{c}{Sensor number} & $8$ & $6$ & $6$ & $16$ & $16$ & $\mathbf{8}$ \\
\hline
Closed-loop & $\times$ & $\surd$ & $\surd$ & $\surd$ & $\surd$ & $\mathbf{\surd}$ \\
\hline
\tabincell{c}{Velocity ($mm/s$)\\in a straight PVC lumen} & - & $5.0$ & $6.0$ & $5.0$ & $6.4$ & $\mathbf{14.1}$ \\
\hline
\tabincell{c}{Velocity ($mm/s$)\\in a curved PVC lumen} & - & $-$ & \tabincell{c}{$5.4$\\($90^{\circ}$ $R200$)} & \tabincell{c}{$4.9$\\($90^{\circ}$ $R200$)} & \tabincell{c}{$4.33$\\($90^{\circ}$ $R200$)} & \tabincell{c}{$\mathbf{10.05}$\\($90^{\circ}$ $R200$) \\ $\mathbf{7.7}$\\($180^{\circ}$ $R18$)} \\
\hline
\tabincell{c}{Working space ($cm^{2}$)} & - & robot's workspace & robot's workspace & $30*30$ & $30*30$ & $\mathbf{60*50}$ \\
\hline
\tabincell{c}{Update Frequency ($Hz$)} & $7$ & $33.3$ & $50$ & $ 0.5 \sim 1.0$ & $25.0$ & $\mathbf{51.3}$ \\
\hline
Position error ($mm$) & $1.7$ & $<5.0$ & $8.5$ & $5.2 \pm 1.0$ & $3.5 \pm 1.4$ & $\mathbf{4.3 \pm 1.9}$ \\
\hline
Orientation error ($^\circ$) & $2.1$ & $<6.0$ & $7.1$ & $5.6 \pm 2.2$ & $9.4 \pm 5.2$ & $\mathbf{5.4 \pm 1.7}$ \\
\hline
\end{tabular}
\label{T_comparison}
\end{table*}

A detailed comparison between our method and previous studies on the permanent magnet-based SMAL systems is listed in Table \ref{T_comparison}. Our method outperforms or performs comparably to the state-of-the-art ones in terms of actuation efficiency (indicated by velocity), workspace size, localization accuracy and update frequency. Besides, due to the usage of the external sensor array, the proposed system design simplifies the inner structure of the capsule, which can reduce the energy consumption and leave more space for other functional modules. Also, our self-adaptive solution achieves the highest localization update frequency and propulsion efficiency (enabled moving speed of the capsule) compared with the state-of-the-art systems.

\subsection{Demonstration videos}

We have provided the detailed demonstration videos of the experiments in Section~V-D to better visualize Fig. \ref{Fig_expdemo}. Specifically, the demonstration of Fig. \ref{Fig_expdemo} (a-d) can be found in the first video\footnote{Available at \url{https://youtu.be/hQwhdF7pRvY} or \url{https://www.bilibili.com/video/BV1Wy4y1y7sF/}} and the demonstration of Fig. \ref{Fig_expdemo} (e) is in the second video\footnote{Available at \url{https://youtu.be/nlbAygg67Dw} or \url{https://www.bilibili.com/video/BV1qK411c7di/}}.
In addition, the failure mode discussed in Fig. \ref{Fig_failure} and Section~V-C is shown in this video\footnote{Available at \url{https://youtu.be/H1_QBZH1BT0} or \url{https://www.bilibili.com/video/BV1Ef4y1x7dv/}}.

\section{Discussion and Conclusion}

In this paper, we present a framework for closed-loop adaptive simultaneous actuation and localization (SMAL) of WCE.
The optimal layout of a fixed number of magnetic sensors is studied and used to select and activate a sensor sub-array in real time to locate the capsule.
A simplified version of the IMOT algorithm is utilized to estimate the 6-D pose of the capsule with improved localization update frequency.
The locomotion of the capsule actuated by a rotating magnetic field in a tubular environment is analyzed, and an adaptive strategy that adjusts the actuator's pose is designed to propel the capsule. The experimental results obtained on the phantoms and ex-vivo pig colons indicate that:
\begin{enumerate}
\item The use of the optimal sensor sub-array with the simplified IMOT algorithm can track the moving capsule with improved accuracy and high update frequency in a large workspace.
\item The real-time adaptive adjustment of the actuator allows efficient propulsion of the capsule in a tubular environment with complex shapes.
\item The overall SMAL system can simultaneously actuate and locate the capsule in a closed-loop manner, showing good robustness in environments with different shapes and materials.
\end{enumerate}

By allowing clinicians to actively control the pose of the capsule, the technology for active WCE locomotion has the potential to improve the diagnostic accuracy and benefit the therapeutic and surgical procedures. With respect to a clinical integration, it should be noted that this work only provides a feasibility study on the active locomotion of the capsule, and future work should clearly focus on the integration of the SMAL modules with the vision, telemetry and power modules in the WCE design \cite{ciuti2011capsule}. 

There are also some limitations of the proposed solution in this work. The optimal sensor layout in the current localization method is designed in 2-D, which would restrict the height of the workspace. Future studies may investigate the 3-D layouts to increase the height of the workspace to accommodate large patients. The velocity-based environment recognition method used in the current system is difficult to determine whether the change in the speed of the capsule is caused by the shape or the resistance of the environment. For instance, if the capsule enters an area of high friction in a straight section, the algorithm may falsely determine the capsule is in a bend and apply less force when more force is probably needed. To this end, the velocity-based environment recognition method can be replaced with a vision-based solution to improve the recognition results based on the real-time images of the intestine \cite{incetan2021vr}\cite{bao2014video}. Also, the current version of the system does not provide an autonomous initialization strategy when the initial heading direction of the capsule is not available. A possible solution to automatic initialization is to automatically change the rotation axis of the actuator to search for the rotation axis of the capsule based on the detection of its rotating status \cite{xu2020novel}. Besides, more complex properties of the human intestinal environment such as peristalsis and deformation are not covered in this paper, which should be carefully modelled in the future work aiming at a clinical integration. Nevertheless, the framework presented in this paper provides a novel solution that can simultaneously actuate and locate a capsule in a tubular environment with convincing performance in an expanded workspace, as well as a potential path for an integration in active WCE to allow more accurate and efficient diagnosis and therapy in the future.


%

\ifCLASSOPTIONcaptionsoff
  \newpage
\fi




\bibliographystyle{IEEEtran}
\bibliography{IEEEabrv,bare_jrnl}

%



%

\begin{IEEEbiography}[{\includegraphics[width=1in,height=1.25in,clip,keepaspectratio]{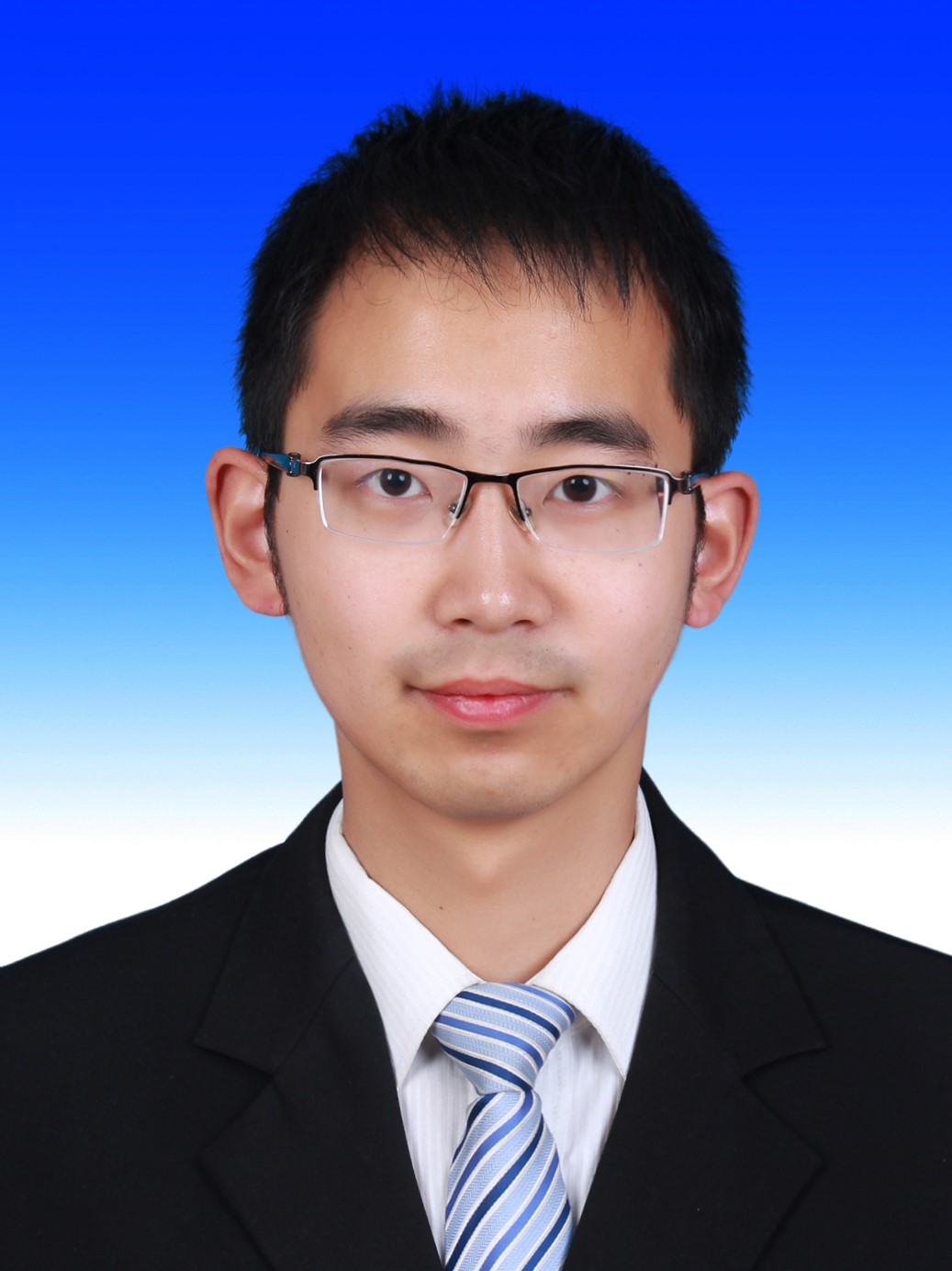}}]{Yangxin Xu}

received the B.Eng. degree in electrical engineering and its automation from Harbin Institute of Technology at Weihai (HIT), Weihai, China, in 2017. He is currently pursuing the Ph.D. degree with the Department of Electronic Engineering, The Chinese University of Hong Kong (CUHK), Hong Kong SAR, China.

His research focuses on magnetic actuation and localization methods and  hardware implementation for wireless robotic capsule endoscopy, supervised by Prof. Max Q.-H, Meng.

Mr. Xu received the Best Conference Paper from the 2018 IEEE International Conference on Robotics and Biomimetics (ROBIO), Kuala Lumpur, Malaysiain, in 2018.

\end{IEEEbiography}

\begin{IEEEbiography}[{\includegraphics[width=1in,height=1.25in,clip,keepaspectratio]{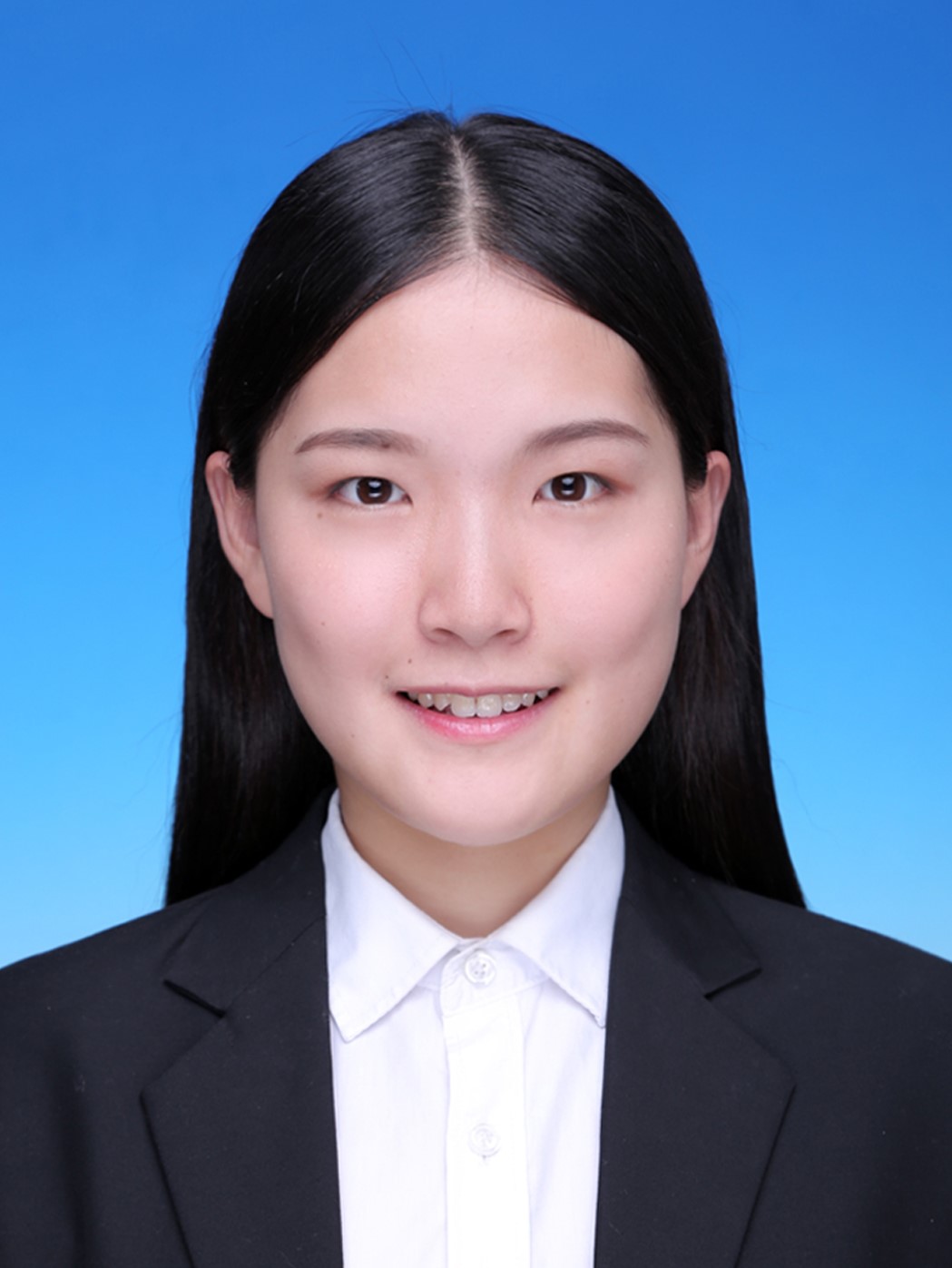}}]
{Keyu Li} received the B.Eng. degree in communication engineering from Harbin Institute of Technology at Weihai (HIT), Weihai, China, in 2019. She is currently pursuing the Ph.D. degree with the Department of Electronic Engineering, The Chinese University of Hong Kong (CUHK), Hong Kong SAR, China.

Her research focuses on medical robots and mobile robot navigation, supervised by Prof. Max Q.-H, Meng.
\end{IEEEbiography}

\begin{IEEEbiography}[{\includegraphics[width=1in,height=1.25in,clip,keepaspectratio]{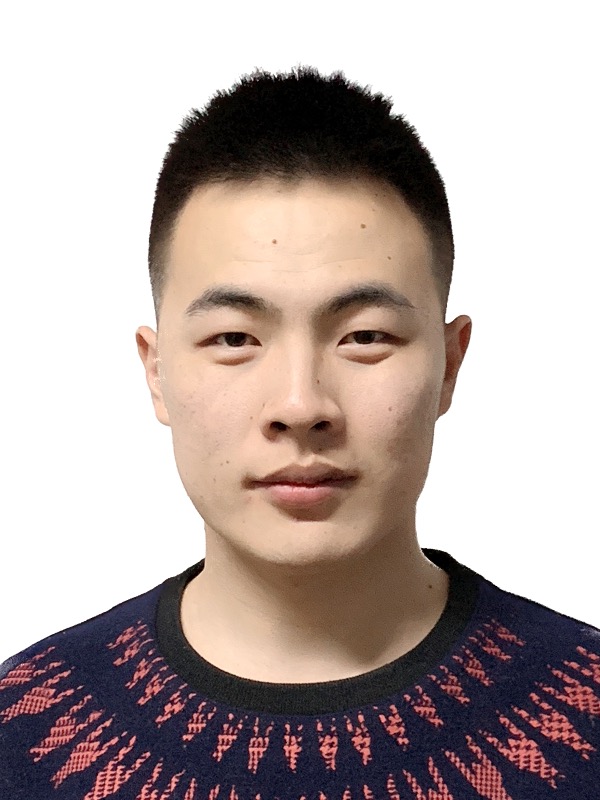}}]
{Ziqi Zhao} recived the B.Eng. and M.Eng. degree in mechanical engineering from Shenyang Jianzhu University, Shenyang, China, in 2016 and 2019, respectively. He is currently pursuing the Ph.D. degree with the Department of Electronic and Electrical Engineering, the Southern University of Science and Technology (SUSTech), Shenzhen, China.

His current research interests include bionic, medical and service robotics, supervised by Prof. Max Q.-H, Meng.

\end{IEEEbiography}

\begin{IEEEbiography}[{\includegraphics[width=1in,height=1.25in,clip,keepaspectratio]{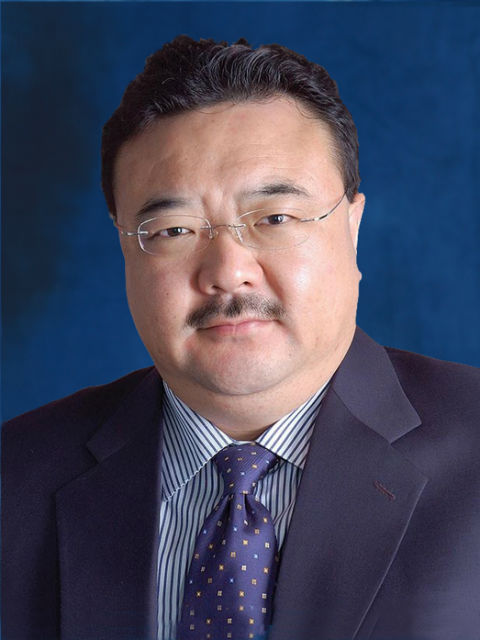}}]
{Max Q.-H. Meng} received the Ph.D. degree in electrical and computer engineering from the University of Victoria, Victoria, BC, Canada, in 1992.

He was with the Department of Electrical and Computer Engineering, University of Alberta, Edmonton, AB, Canada, where he served as the Director of the Advanced Robotics and Teleoperation Laboratory, holding the positions of Assistant Professor, Associate Professor, and Professor in 1994, 1998, and 2000, respectively. In 2001, he joined The Chinese University of Hong Kong, where he served as the Chairman of the Department of Electronic Engineering, holding the position of Professor. He is affiliated with the State Key Laboratory of Robotics and Systems, Harbin Institute of Technology, and is the Honorary Dean of the School of Control Science and Engineering, Shandong University, China. He is currently with the Department of Electronic and Electrical Engineering, Southern University of Science and Technology, on leave from the Department of Electronic Engineering, The Chinese University of Hong Kong, Hong Kong SAR, China, and also with the Shenzhen Research Institute of the Chinese University of Hong Kong, Shenzhen, China. His research interests include robotics, medical robotics and devices, perception, and scenario intelligence. He has published about 600 journal and conference papers and led more than 50 funded research projects to completion as PI.

Dr. Meng is an elected member of the Administrative Committee (AdCom) of the IEEE Robotics and Automation Society. He is a recipient of the IEEE Millennium Medal, a fellow of the Canadian Academy of Engineering, and a fellow of HKIE. He has served as an editor for several journals and also as the General and Program Chair for many conferences, including the General Chair of IROS 2005 and the General Chair of ICRA 2021 to be held in Xi'an, China.
\end{IEEEbiography}







\end{document}